\documentclass{article}

\usepackage{microtype}
\usepackage{graphicx}
\usepackage{subfigure}
\usepackage{booktabs} % for professional tables

\usepackage{hyperref}

\usepackage[accepted]{icml2025}

\usepackage{amsmath}
\usepackage{amssymb}
\usepackage{mathtools}
\usepackage{amsthm}
\usepackage{capt-of}

\usepackage[capitalize,noabbrev]{cleveref}

\theoremstyle{plain}

\theoremstyle{definition}

\theoremstyle{remark}

\usepackage[textsize=tiny]{todonotes}

\icmltitlerunning{DragLoRA: Online Optimization of LoRA Adapters for Drag-based Image Editing in Diffusion Model
}

\begin{document}

\twocolumn[
\icmltitle{DragLoRA: Online Optimization of LoRA Adapters for Drag-based Image Editing in Diffusion Model}

%\begin{figure*}

%\end{figure*}
% It is OKAY to include author information, even for blind
% submissions: the style file will automatically remove it for you
% unless you've provided the [accepted] option to the icml2025
% package.

% List of affiliations: The first argument should be a (short)
% identifier you will use later to specify author affiliations
% Academic affiliations should list Department, University, City, Region, Country
% Industry affiliations should list Company, City, Region, Country

% You can specify symbols, otherwise they are numbered in order.
% Ideally, you should not use this facility. Affiliations will be numbered
% in order of appearance and this is the preferred way.
\icmlsetsymbol{equal}{*}

\begin{icmlauthorlist}
\icmlauthor{Siwei Xia}{yyy}
\icmlauthor{Li Sun}{yyy,2}
\icmlauthor{Tiantian Sun}{yyy}
\icmlauthor{Qingli Li}{yyy}
\end{icmlauthorlist}

\icmlaffiliation{yyy}{Shanghai Key Laboratory of Multidimensional Information Processing.}
\icmlaffiliation{2}{Key Laboratory of Advanced Theory and Application in Statistics and Data Science,  East China Normal University, Shanghai, China}
% \icmlaffiliation{sch}{School of ZZZ, Institute of WWW, Location, Country}

\icmlcorrespondingauthor{Li Sun}{sunli@ee.ecnu.edu.cn}
% \icmlcorrespondingauthor{Firstname2 Lastname2}{first2.last2@www.uk}

% You may provide any keywords that you
% find helpful for describing your paper; these are used to populate
% the "keywords" metadata in the PDF but will not be shown in the document
\icmlkeywords{Generative Models, Image Editing, Stable Diffusion.}

\begin{center}
\centerline{\includegraphics[width=\textwidth]{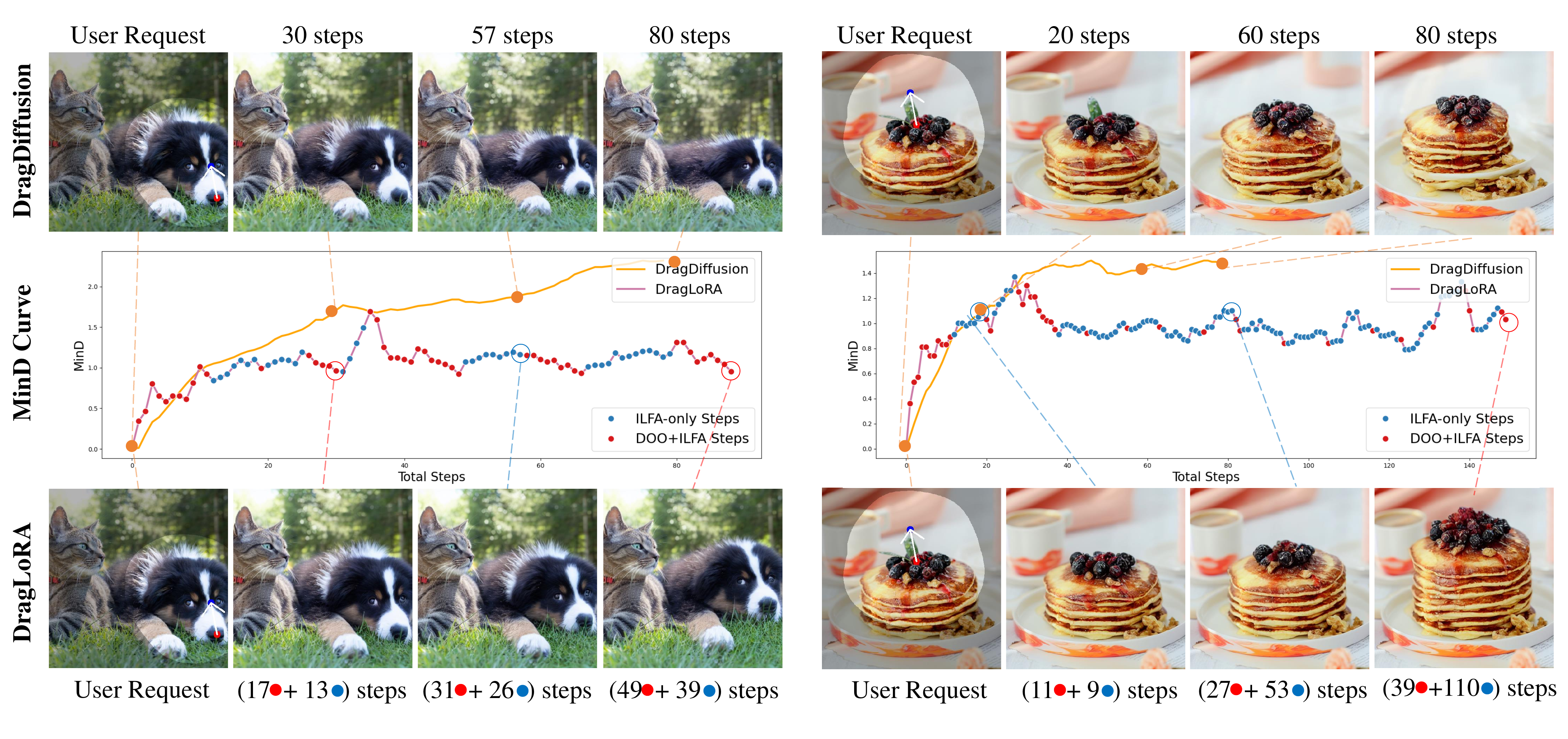}}
\captionof{figure}{%Visual comparison between DragDiffusion \cite{shi2024dragdiffusion} and the proposed DragLoRA at each optimization step. For a given image and user's request, dragged images at four intermediate steps are provided. On the top, DragDiffusion takes the full (80) steps but gives generation with less fidelity. On the bottom, DragLoRA takes less steps to drive the source point to the target. Meanwhile, its results are more accurate with. In the middle, $minD$ curves are visualized, showing that the tracked points from DragLoRA are more similar to the source points than DragDiffusion. 
Visual comparison between DragDiffusion \cite{shi2024dragdiffusion} and DragLoRA at each step. For a given image and user request, we present dragged images at four intermediate steps. 
In DragLoRA, the steps are divided into two types: DOO+ILFA (Red) or ILFA-only (Blue), where DOO stands for Dual-Objective Optimization and ILFA stands for Input Latent Feature Adaptation.
DragDiffusion
requires 80 optimization steps but produces results with lower fidelity. In contrast, DragLoRA
achieves more precise deformations with fewer optimization steps and more total steps, which consumes less time thanks to the high-efficiency of ILFA.  
The middle \( minD \) curves demonstrate that DragLoRA achieves better point tracking, preserving source details more accurately than DragDiffusion.}
\label{fig：teaser}
\end{center}
%\vskip -0.2in
% \vskip 0.3in
]

% this must go after the closing bracket ] following \twocolumn[ ...

% This command actually creates the footnote in the first column
% listing the affiliations and the copyright notice.
% The command takes one argument, which is text to display at the start of the footnote.
% The \icmlEqualContribution command is standard text for equal contribution.
% Remove it (just {}) if you do not need this facility.

\printAffiliationsAndNotice{}  % leave blank if no need to mention equal contribution
% \printAffiliationsAndNotice{\icmlEqualContribution} % otherwise use the standard text.

\begin{abstract}
Drag-based editing within pretrained diffusion model provides a precise and flexible way to manipulate foreground objects. Traditional methods optimize the input feature obtained from DDIM inversion directly, adjusting them iteratively to guide handle points towards target locations. However, these approaches often suffer from limited accuracy due to the low representation ability of the feature in motion supervision, as well as inefficiencies caused by the large search space required for point tracking. To address these limitations, we present DragLoRA, a novel framework that integrates LoRA (Low-Rank Adaptation) adapters into the drag-based editing pipeline. To enhance the training of LoRA adapters, we introduce an additional denoising score distillation loss which regularizes the online model by aligning its output with that of the original model. Additionally, we improve the consistency of motion supervision by adapting the input features using the updated LoRA, giving a more stable and accurate input feature for subsequent operations. Building on this, we design an adaptive optimization scheme that dynamically toggles between two modes, prioritizing efficiency without compromising precision. Extensive experiments demonstrate that DragLoRA significantly enhances the control precision and computational efficiency for drag-based image editing. %, reducing both the optimization steps and the search space for handle points. %Experiments demonstrate that DragLoRA, with the reduced optimization step and the narrow search space of handle points, significantly improves the editing precision and the computational efficiency for drag-based image editing. %and we also adapt the input feature for the next step motion supervision by denoising and re-noising using the updated LoRA, ensuring consistent and efficient updates.
The Codes %of DragLoRA 
are available at: https://github.com/Sylvie-X/DragLoRA.
\end{abstract}

%begin{figure*}[ht]
% \vskip 0.2in
%\begin{center}
% \centerline{\includegraphics[width=\columnwidth]{drag_overall}}
%\centerline{\includegraphics[width=(\columnwidth)*2]{first}}
%\caption{first}
%\label{drag-first}
%\end{center}
%\vskip -0.2in
%\end{figure*}

\section{Introduction}
\label{sec:intro}
%Stable Diffusion (SD) model is able to synthesize high-quality images based on input text prompts. 
The Stable Diffusion (SD) model has demonstrated remarkable capabilities in synthesizing high-quality images from textual prompts. Based on the pre-trained SD model, many works \cite{hertz2022prompt,zhang2023adding,hertz2023delta} have sought to enhance control over the %content of 
generated images, often relying on detailed text prompts or reference images to specify generation conditions. %They rely on either input text prompts or reference images to specify the condition for generation. 
While these methods offer various types of editing, they typically require users to provide complex instructions to describe their desired outputs, which can be cumbersome and restrictive in scenarios where only minimal modification is required. %However, these methods require the user to provide complex input to describe their requirement, which limits their applications for those settings only small .

Recent advances in drag-based image editing enable intuitive point-driven manipulation within pre-trained generative models. By specifying pairs of source and target points, users can interactively guide object deformation, iteratively "dragging" content from the source location toward the target. These methods typically operate through two sequential stages: motion supervision, which computes directional gradients to align features with the desired movement, and point tracking, which updates handle positions based on the evolving feature space. 
While this paradigm reduces reliance on complex textual or reference inputs, existing approaches often struggle with precision and efficiency. Direct optimization of latent features, such as DDIM-inverted representations, introduces instability due to limited feature expressiveness in motion supervision, while the iterative tracking process incurs high computational costs from searching large spatial regions. 

%In this work, we address these limitations by enhancing the diffusion model with an added LoRA adapter, inserted in all attention layers of the pre-trained SD model and optimized in an online manner for drag-based editing. Equipped with more parameters in the adapter, the proposed DragLoRA is expected to have a stronger ability to represent the foreground object, hence is able to deform the handle point into the specified position accurately, therefore, increases the accuracy in every step of motion supervision. Since 
In this work, we address these limitations by augmenting the SD model with DragLoRA, a dynamically optimized adapter integrated into all attention layers of the pre-trained Unet. Unlike prior methods that directly optimize latent features, our method performs online adaptive learning during drag-based editing, enabling adjustments of LoRA parameters tailored to user interactions. The expanded optimization space enhances the model’s capacity to represent foreground deformations, allowing precise alignment of handle points with target positions at each motion supervision step. By decoupling deformation control from static latent representations, DragLoRA mitigates the instability caused by limited feature expressiveness, hence reducing reliance on iterative large-scale searches. 

However, we observe that unrestricted LoRA optimization guided solely by drag loss can lead to deviations from the original image. To address this, we introduce a delta denoising score (DDS) loss to regularize the online training. %Specifically, we utilize the DragLoRA Unet to make a single-step prediction for the clean feature based on the input feature from the DDIM inversion, perturb it with noise at a randomly sampled timestep, and then compute the difference between the noise predictions of the original and LoRA-enhanced UNets. 
Specifically, DragLoRA first predicts a clean feature from the DDIM-inverted input at timestep of $t$, which is then perturbed with noise at a randomly sampled timestep $t'$. The DDS loss is computed as the difference between the noise predictions of the original UNet and the DragLoRA enhanced UNet for the perturbed feature. %By minimizing this distillation loss alongside the drag objective, we enforce fidelity to the source image while retaining deformation flexibility.  
By jointly minimizing the extra loss with the original motion supervision objective, our method preserves semantic fidelity %to the source image 
while enabling flexible deformations.
Furthermore, to ensure motion consistency across iterative edits, we adapt the input feature to accumulate deformation effects. At each step, the input feature is denoised from $t$ to 
$t-1$ using DragLoRA’s predicted noise, then re-noised back to 
$t$ with random perturbations, for the next optimization. 
This cycle progressively aligns the feature with the accumulated deformation trajectory, propagating handle point adjustments into the latent space and stabilizing motion supervision through coherent feature updates. 

In practice, %we find that handle points can be driven towards target points only by the input adaptation but without motion supervision. 
we observe that handle points can be driven toward target positions through input adaptation alone, even without explicit motion supervision. This occurs because accumulated gradients from previous optimizations can be utilized for moving handle points at the new positions without extra driving force.
In each gradient step, although the specific tasks are not exactly the same, they share a low-variance handle feature and a common direction. DragLoRA can learn these commonalities and generalize, which is comparable with meta-learning.
% DragLoRA learns to drag and it inherently deforms and update the input feature. 
%Therefore, DragLoRA can be adaptively optimized according to the quality of point tracking. DragLoRA is able to narrow the search space for point tracking, enabling efficient handle localization without sacrificing deformation accuracy.
To leverage this, we employ an adaptive optimization strategy: when point tracking achieves sufficient quality, %—measured by the feature similarity between tracked and original points, and the proximity of tracked points to interim targets—
LoRA updates are bypassed to prioritize efficiency. Conversely, if tracking deviates (e.g., due to occlusions or ambiguous textures), motion supervision is triggered to refine the LoRA parameters, ensuring robust deformation control. By dynamically toggling between motion supervision and input adaptation, DragLoRA %narrows the search space for point tracking while maintaining accuracy. This dual mechanism 
enables efficient handle localization with minimal optimization steps, as it selectively optimizes LoRA only when necessary.
%Moreover, the input to DragLoRA also needs to be adapted so that the motion at the handle point is accumulated at the input image feature, and this ensures the consistency of motion supervision. Particularly, we use the predicted noise from DragLoRA and denoise the input feature from $t$ to $t-1$. Then random noise is added onto the denoised feature, changing it back to the step $t$ again for future motion supervision.
%Therefore, it not only improves per-step accuracy but also streamlines the tracking process, realizing a fast and accurate user-guided editing. %However, in practice, we find that DragLoRA needs more constraint during motion supervision to maintain consistency with the original image. Here, a score distillation sampling loss is incorporated for motion supervision. In particular, the predicted clean image is first computed from the Unet with LoRA. Then, noises are added to it in a random time step. The perturbed image is then given to both original and DragLoRA Unet, computing the difference between them for training the LoRA. More trainable parameters tend to
%Recently, drag-based methods realize interactive point-based editing within the pre-trained generation model. The user provides the source and target positions for image editing, and the method iteratively moves the contents of the source point to the corresponding target. Typically, such methods have two types of operations, motion supervision and point tracking. The former.

The contributions of this paper lie in following aspects.
\begin{itemize}
\item We propose DragLoRA, a parameterized adapter enables online optimization following user's interactions. By replacing direct latent feature optimization with dynamic model adaptation, DragLoRA enhances fine-grained deformation capacity while preserving the pretrained diffusion priors. %It replaces direct latent feature optimization, expanding the model’s capacity for fine-grained deformations while maintaining pretrained diffusion priors. 
\item We introduce a dual-objective framework that combines drag loss with a DDS loss, computed by comparing noise predictions of the original and LoRA-augmented Unets on perturbed features. Coupled with a cyclic denoise-renoise process, which iteratively propagates handle point adjustments into the latent space, this framework ensures semantic fidelity with the source image and stabilizes motion supervision through accumulated deformation trajectories. %Meanwhile, we propose a cyclic denoise-renoise process to adapt input features iteratively: denoising with DragLoRA’s predictions, then renoising to propagate handle point adjustments into the latent space. These two schemes ensure semantic fidelity to the source image and stabilize the motion supervision with accumulated deformation trajectories. 
\item We design an adaptive optimization strategy that dynamically toggles between two modes. %input feature adaptation and LoRA parameter updates. 
By evaluating tracking quality, DragLoRA prioritizes efficient input feature updates when tracking succeeds and activates motion supervision for LoRA refinement when deviations occur, minimizing redundant optimization steps. 
\end{itemize}

\section{Related Works}
\textbf{Applications of diffusion model on image editing.} Diffusion models \cite{ho2020denoising, song2020score, dhariwal2021diffusion} are initially designed for iterative denoising in the pixel domain and later accelerated by operating in the latent feature space \cite{rombach2022high}. Trained on large-scale image-text datasets, Stable Diffusion can generate high-quality images conditioned on input text. These models enable various image editing tasks with little to no additional training. Typically, the source image is noised to an intermediate timestep suitable for editing, then denoised back with target conditions to modify the content. SDEdit \cite{meng2022sdedit} applies random noise directly following the DDPM schedule, while DDIM inversion \cite{song2020denoising} and its advanced versions \cite{mokady2023null,miyake2023negative} have been shown to better preserve details from the source image. 

Among various editing techniques, text-based editing is one of the most explored. P2P \cite{hertz2022prompt} achieves precise modifications by adjusting the cross-attention matrix in the UNet, while PnP \cite{tumanyan2023plug} and Masactrl \cite{cao2023masactrl} focus on modifying the self-attention layers. Optimization-based methods, on the other hand, are typically trained on a single image \cite{kawar2023imagic, valevski2023unitune, hertz2023delta} or a small set of images \cite{gal2022image, ruiz2023dreambooth, kumari2023multi}, allowing the model to learn detailed appearances of the editing target and adapt to arbitrary text prompts. In contrast, reference-based editing directly feeds a reference image into the model, often requiring a large dataset for effective learning \cite{wei2023elite, ye2023ip}. The reference can take various forms, including a standard image or structural information extracted from the target image \cite{zhang2023adding}. %P2P \cite{hertz2022prompt} enables precise modifications by adjusting the cross-attention matrix in the Unet, while PnP \cite{tumanyan2023plug} and Masactrl \cite{cao2023masactrl} focus on modifying the self-attention layer. Other optimization-based methods are typically trained for a single \cite{kawar2023imagic,valevski2023unitune,hertz2023delta} or a few images \cite{gal2022image,ruiz2023dreambooth,kumari2023multi}, so the model knows appearances of editing object well and it can be adapted with arbitrary target text prompt. On the other hand, reference-based editing directly input the reference to the model, and it usually needs a large dataset \cite{wei2023elite,ye2023ip}. The reference can be in various forms, such as common image, or reflecting the structure of the image \cite{zhang2023adding}.

\textbf{Dragged-based image editing} 
has gained significant attention for its intuitive approach to modifying images through user-defined handle and target points. DragGAN \cite{pan2023drag} is the first to demonstrate the feasibility of "dragging" in pre-trained StyleGAN models, introducing motion supervision and point tracking mechanisms. Subsequent advancements have expanded this concept to diffusion models. DragDiffusion \cite{shi2024dragdiffusion} adapts the point-based dragging technique to the SD model, enhancing generative quality and realizing precise spatial control.
SDE-Drag \cite{nie2024blessing} presents a unified probabilistic formulation for diffusion-based image editing, including dragging. 
DragNoise \cite{liu2024dragnoise} utilizes U-Net’s noise predictions for efficient point-based editing while maintaining semantic coherence. FreeDrag \cite{ling2024freedrag} improves stability and efficiency by narrowing the search region for handle points and incorporating adaptive feature updates. EasyDrag \cite{hou2024easydrag} simplifies the user interaction process, making image editing more intuitive and accessible. GoodDrag \cite{zhang2024gooddrag} introduces an alternating drag and denoising framework, improving result fidelity and reducing distortion. StableDrag \cite{cui2024stabledrag} addresses challenges in point tracking and motion supervision by developing a discriminative point tracking method and a confidence-based latent enhancement strategy, resulting in more stable and precise drag-based editing. AdaptiveDrag \cite{chen2024adaptivedrag} proposes a mask-free point-based editing approach, leveraging super-pixel segmentation for adaptive steps. %leading to more flexible and accurate editing outcomes, 
ClipDrag \cite{jiang2024clipdrag} leverages CLIP for text-guided editing, offering semantic control over image content, while DragText \cite{choi2024dragtext} facilitates text-guided drag by optimizing text embeddings alongside image features. FastDrag \cite{zhao2024fastdrag} enabled quick image modifications without the need for iterative optimization. GDrag \cite{lin2025gdrag} categorizes point-based manipulations into three atomic tasks with dense trajectory, achieving less ambiguous outputs.%RegionDrag further enhanced user control by allowing editing instructions in the form of handle and target regions, enabling more precise manipulation and faster processing times.

%Beside point-based drag editing, %On the optimization-free front, 
%DragonDiffusion \cite{mou2023dragondiffusion}, DiffEditor \cite{mou2024diffeditor} and RegionDrag \cite{lu2024regiondrag} extend the drag-based editing to regions or introducing a reference for editing. %directly move region. 
%Additionally, InstantDrag \cite{shin2024instantdrag} and LightningDrag \cite{shi2024lightningdrag} train general-purpose models for drag-based editing, allowing for rapid adaptation to various tasks and datasets.

Besides point-based drag editing, DragonDiffusion \cite{mou2023dragondiffusion}, DiffEditor \cite{mou2024diffeditor}, and RegionDrag \cite{lu2024regiondrag} extend drag editing to regions or incorporate a reference image for editing. Additionally, InstantDrag \cite{shin2024instantdrag} and LightningDrag \cite{shi2024lightningdrag} train general-purpose models for drag-based editing, enabling rapid adaptation across various tasks and datasets.

The proposed DragLoRA is a point-based drag editing method. Compared to other methods in the same category, it achieves state-of-the-art performance on key metrics while reducing time costs. In contrast to general models like LightningDrag, DragLoRA avoids the heavy burden of offline training and delivers better results. %These developments collectively advance the field of drag-based image editing, offering improved stability, flexibility, and user-friendliness across various applications. 
%DragGAN \cite{pan2023drag} is the first work showing that "“Drag” is possible in pre-trained StyleGAN. It proposes the motion supervision and point tracking. %Drag editing aims to change the position, pose or shape of the foreground object directly according to user. 

\begin{figure*}[ht]
% \vskip 0.2in
\begin{center}
\centerline{\includegraphics[width=(\columnwidth)*2]{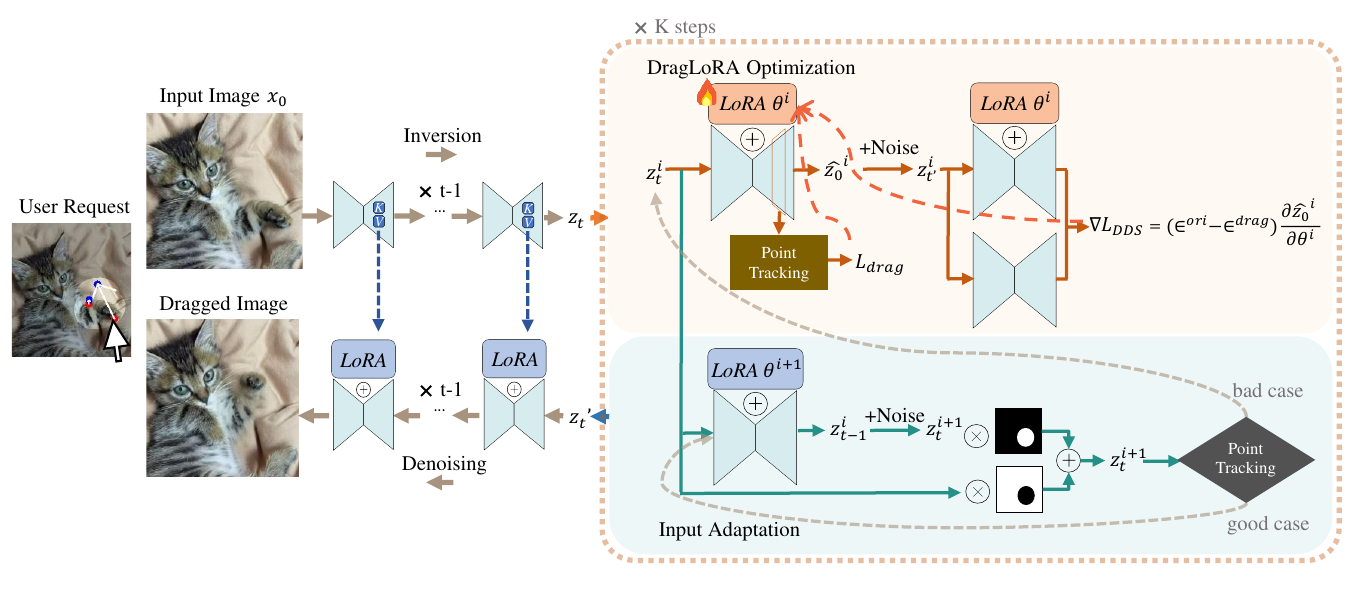}}
\caption{%Overview of our proposed DragLoRA. Given a DDIM inversion code $z_t$ with $t=35$ from a source image $x_0$, we incorporate a LoRA adapter and train it based on $L_\text{drag}$ and $L_\text{DDS}$ in an online optimization manner. $L_\text{drag}$ is a major guidance for driving handle points, while $L_\text{DDS}$ imposes an extra constraint on LoRA to keep the model not being far away from the original one, hence increasing the fidelity of the dragged image. The input feature to the Unet is also adapted by a denoise-renoise cycle and it can drive a handle point by LoRA-adapter without optimization gradient. DragLoRA switches between the common motion supervision and input-adaptation-only modes based on the quality from point tracking.
Overview of our proposed DragLoRA. Given an inversion code \( z_t \) at \( t = 35 \) from a source image \( x_0 \), we incorporate a LoRA adapter and optimize it online using \( L_\text{drag} \) and \( L_\text{DDS} \). \( L_\text{drag} \) primarily guides handle point movement, while \( L_\text{DDS} \) constrains LoRA to remain close to the original model, preserving the fidelity of the edited image. Additionally, the input feature to the UNet undergoes a denoise-renoise cycle in the foreground region, allowing the LoRA adapter to drive handle points even without gradient updates. DragLoRA dynamically switches between motion supervision and input-adaptation-only modes based on point tracking quality, ensuring both efficiency and stability in the editing process. }
\label{drag-overall}
\end{center}
\vskip -0.1in
\end{figure*}

\section{Method}
The proposed DragLoRA framework, as is shown in Figure \ref{drag-overall}, augments the pretrained Stable Diffusion (SD) model with a dynamically optimized adapter to enable interactive drag-based image editing. Our method has two modes for driving handle points, which are LoRA training with motion supervision and input feature adaptation without it. And we design an adaptive scheme to combine them for drag-based editing.%Following the pipeline of DragDiffusion \cite{dragdiffusion}, we first fine-tune the UNet using a LoRA specific to the input image, then apply DDIM inversion to obtain a deterministic latent trajectory for motion supervision at timestep $t$. However, unlike prior work that directly optimizes latent features, our method introduces three key innovations: (1) online LoRA adaptation to decouple deformation control from static latents, (2) delta denoising score (DDS) loss to preserve semantic consistency, and (3) cyclic feature propagation to stabilize iterative motion supervision. These components work synergistically to narrow the optimization search space while improving precision—enabling efficient handle point tracking without sacrificing edit flexibility.

\subsection{Preliminaries: Diffusion Model and Drag-based Image Editing}
%Diffusion models \cite{ho2020denoising, rombach2022high} establish a probabilistic framework for generating images through iterative denoising. In the forward process, a clean signal $z_0$, typically in pixel or a compressed latent space, is gradually corrupted by additive Gaussian noise over $T$ steps. This process can be defined as:
%\begin{equation}\label{eq:eq1}
% z_t = \sqrt{\bar{\alpha}_t} \cdot z_0 + \sqrt{1 - \bar{\alpha}_t} \cdot \epsilon, \quad \epsilon \sim \mathcal{N}(0,I)
%\end{equation}
%where $\bar{\alpha}_t = \prod_{s=1}^t (1-\beta_s)$ denotes the cumulative product of noise retention coefficients, and $\{\beta_t\}$ forms a predefined variance schedule controlling the noise injection rate. As $t$ increases, $\bar{\alpha}_t$ monotonically decreases, progressively transforming the original signal into isotropic Gaussian noise. Enhanced by text conditions and a large number of training samples, the SD model can give high-quality generations by predicting $\epsilon$ with Unet parameterized by $\epsilon_\theta ()$.
\textbf{Diffusion models} \cite{ho2020denoising, rombach2022high} %synthesize images through iterative denoising of noise-corrupted signals. In 
have the forward process, in which an input image or latent representation \( z_0 \) is progressively perturbed by Gaussian noise over \( T \) timesteps:   
\begin{equation}\label{eq:eq1}
\begin{split}
z_t &= \sqrt{\alpha_t} \, z_{t-1} + \sqrt{1 - \alpha_t} \, \epsilon_t \\
&= \sqrt{\bar{\alpha}_t} \, z_0 + \sqrt{1 - \bar{\alpha}_t} \, \epsilon, \quad \epsilon_t, \epsilon \sim \mathcal{N}(0,I),
\end{split}
\end{equation}
where \( \alpha_t=1-\beta_t\) and \( \bar{\alpha}_t = \prod_{s=1}^t (1-\beta_s) \) is the cumulative noise retention coefficient, and \( \{\beta_t\} \) defines a fixed noise schedule. As \( t \to T \), \( \bar{\alpha}_t \) decays to zero, and \( z_t \) becomes a pure noise.  
The reverse process aims to recover \( z_0 \) by iteratively denoising \( z_t \). Denoising Diffusion Implicit Models (DDIM) \cite{song2020denoising} propose a non-Markovian sampling process:  
\begin{equation}\small\label{eq:eq2} 
z_{t-1} = \sqrt{\bar{\alpha}_{t-1}} \left( \frac{z_t - \sqrt{1 - \bar{\alpha}_t} \, \epsilon_\theta(z_t, t)}{\sqrt{\bar{\alpha}_t}} \right) + \sqrt{1 - \bar{\alpha}_{t-1}} \cdot \epsilon_\theta(z_t, t),
\end{equation}   
giving an accelerated generation. Crucially, DDIM supports deterministic inversion: given \( z_0 \), the noisy latent \( z_t \) at any \( t \) can be reconstructed via:  
\begin{equation}\small\label{eq:eqddim}
z_{t+1} = \sqrt{\bar{\alpha}_{t+1}} \left( \frac{z_t - \sqrt{1 - \bar{\alpha}_t} \, \epsilon_\theta(z_t, t)}{\sqrt{\bar{\alpha}_t}} \right) + \sqrt{1 - \bar{\alpha}_{t+1}} \, \epsilon_\theta(z_t, t).
\end{equation} 
This inversion maps \( z_0 \) to an editable latent at time $t$ %(e.g., \( z_{35} \) at \( t=35 \)) 
for different tasks, \emph{e.g.}, \( z_{35} \) for drag editing. %serving as the optimization anchor for . %optimized by motion supervision in drag-based editing. 

%DragGAN is the first work that proposes drag-based image editing. DragDiffusion \cite{dragdiffusion} is the first work that support drag-based image editing . realizes interactive image editing by optimizing 
\textbf{Drag-based image editing} is typically performed in a pretrained generative model, such as GAN or diffusion-based model. In this task, user provides a set of source points $\mathbf{p}_i$ and corresponding target points $\mathbf{g}_i$, where $\mathbf{p}_i,\mathbf{g}_i\in \mathbb{R}^2$ 
represent
2D pixel coordinates in the image plane. For SD, a LoRA adapter is first fine-tuned on the input image to ensure that the edited results retain a high similarity to the original. The latent feature,  \( z_{35}^0 \) at timestep \( t=35 \), is then obtained via DDIM inversion and serves as the optimization target. During editing, two stages alternate: 

\textbf{Motion Supervision}: A gradient-based objective adjusts \( z_{35} \) 
to morph the neighborhood of temporal target points $\mathbf{h}_i+\mathbf{d}_i$ into the corresponding regions around handle points $\mathbf{h}_i$, 
% to align handle points with target positions, 
using features from a selected UNet layer to compute directional guidance, as is shown in (\ref{eq:eq3}).  
\begin{equation}\label{eq:eq3} 
L_\text{drag}=\sum_{i=1}^{N} \| \mathrm{sg}(F(z_{35}^0, \mathbf{h}_i^0)) - F(z_{35}, \mathbf{h}_i+\mathbf{d}_i) \|_1
% L_\text{drag}=\sum_{i=1}^{N} \| \mathrm{sg}(F(z_{35}, \mathbf{h}_i)) - F(z_{35}, \mathbf{h}_i+\mathbf{d}_i) \|_1
\end{equation}  
where \( F(\cdot) \) extracts features from the specified UNet layer and \(\mathrm{sg}(\cdot)\) detaches the possible gradients. $N$ denotes total number of handle points specified by user.  \( \mathbf{d}_i \) is the normalized displacement vectors $
\mathbf{d}_i=\frac{\mathbf{g}_i-\mathbf{h}_i}{\|\mathbf{g}_i-\mathbf{h}_i\|_2}
$. To mitigate cumulative errors, we adopt the initial latent input \( z_{35}^0\) and handle point \(\mathbf{h}_i^0 =\mathbf{p}_i\) to obtain fixed target features, which is different from \cite{pan2023drag,shi2024dragdiffusion}. %$L_\text{drag}$ aligns handle coordinates \( \mathbf{h} \) with temporal target coordinates \( \mathbf{h+d} \). 
To preserve the original content outside the edited regions, an optional constraint \( L_{\text{Mask}} \) is applied:  
\begin{equation}\label{eq:eq4}  
L_{\text{Mask}} = \| (z_{34} - z_{34}^0) \cdot (1 - M) \|_1,  
\end{equation}  
where \( z_{34} \) is denoised from \(z_{35}\) by (\ref{eq:eq2}) through a forward pass of the model, and \( M \) is a given binary mask indicating editable regions. This ensures that non-target areas remain unchanged during optimization. 

\textbf{Point Tracking}: The updated \( \hat{z}_{35} \) is reprocessed through the UNet to locate the new handle positions \( \mathbf{h}_i^{k+1} \), 
which serve as new guidance for subsequent motion supervision.
% iteratively refining the deformation.
Here, \( \Omega(\mathbf{h}_i^k, r_2) \) defines a rectangular search area centered at the previous handle point \( \mathbf{h}_i^k \), with \( r_2 \) controlling its size.

\begin{equation}\label{eq:eq6}
\mathbf{h}_i^{k+1} = \arg \min_{\mathbf{q} \in \Omega(\mathbf{h}_i^k, r_2)} \left\| F(\hat{z}_{35}, \mathbf{q}) - F(z_{35}^0, \mathbf{h}_i^0) \right\|_1
\end{equation}

Additionally, the new handle point \( \mathbf{h}_i^{k+1} \) can be used to assess the quality of drag. We compute the best matching distance between \( \mathbf{h}_i^{k+1} \) and \( \mathbf{h}_i^0 \), as defined in (\ref{eq:eq7}). A lower value of \( minD \) indicates higher confidence in point tracking and the success of the previous optimization for motion supervision.

\begin{equation}\label{eq:eq7}
min D = \left\| F(\hat{z}_{35}, \mathbf{h}_i^{k+1}) - F(z_{35}^0, \mathbf{h}_i^0) \right\|_1
\end{equation}

Furthermore, we evaluate a geometric metric, the Euclidean distance \( d(\mathbf{h}_i^{k+1}, \mathbf{h}_i^k + \mathbf{d}_i) \), between \( \mathbf{h}_i^{k+1} \) and \( \mathbf{h}_i^k + \mathbf{d}_i \). Intuitively, this distance should be small, as motion supervision requires \( F(z_{35}, \mathbf{h}_i^k + \mathbf{d}_i) \) to align closely with \( F(z_{35}^0, \mathbf{h}_i^0) \) from the previous optimization step. Thus, the tracked point should not deviate significantly from it. These additional metrics are not used in previous methods and are specific to our proposed online adaptive optimization strategy. Further details can be found in Section \ref{ass}.

%\textbf{Point Tracking}: The updated \( \hat{z}_{35} \) is reprocessed through the UNet to locate new handle positions $\mathbf{h}_i^{k+1}$ on the feature map, iteratively refining the deformation. Here $\Omega(\mathbf{h}_i^k, r_2)$ is a rectangle searching area centered at the previous handle point $\mathbf{h}_i^k$. $r_2$ controls the size of it.  
%\begin{equation}\label{eq:eq6}
%\mathbf{h}_i^{k+1} = \arg \min_{\mathbf{q} \in \Omega(\mathbf{h}_i^k, r_2)} \left\| F(\hat{z}_{35},\mathbf{q}) - F(z_{35}^0,\mathbf{h}_i^0) \right\|_1
%\end{equation}
%Moreover, we find the new handle point $\mathbf{h}_i^{k+1}$ can be used to evaluate the quality of drag. We are interested in calculating the best matching distance between $\mathbf{h}_i^{k+1}$ and $\mathbf{h}_i^{0}$ which is defined as (\ref{eq:eq7}). The small value of $minD$ denotes a high confidence of point tracking and previous optimization for motion supervision.
%\begin{equation}\label{eq:eq7}
%min D =\left\| F(\hat{z}_{35},\mathbf{h}_i^{k+1}) - F(z_{35}^0,\mathbf{h}_i^0) \right\|_1
%\end{equation}
%Besides, we evaluate a geometry metric which is the Euclidean distance $d(\mathbf{h}_i^{k+1},\mathbf{h}_i^k+\mathbf{d}_i)$ between $\mathbf{h}_i^{k+1}$ and $\mathbf{h}_i^k+\mathbf{d}_i$. Intuitively, this metric should be small enough since the motion supervision requires $F({z}_{35},\mathbf{h}_i^k+\mathbf{d}_i)$ to be close with $F({z}_{35}^0,\mathbf{h}_i^0)$ in previous optimization step. Thus the tracked point is not far away from it.

\subsection{DragLoRA and Its Online Optimization}
\label{doo}
Building on existing works, we introduce DragLoRA, a novel framework that improves the precision and efficiency of user-guided deformations through online optimization of LoRA adapters. Instead of directly optimizing latent feature \( z_{35} \), DragLoRA dynamically adjusts the parameters $\Delta \theta$ of LoRA, which are integrated into the UNet parameterized by $\theta$. This approach increases the capacity of the model and decouples deformation control from latent features, giving fine-grained adjustments while preserving semantic fidelity. To minimize additional computation, we initialize the LoRA with the weights from reconstruction fine-tuning on the input image, without introducing extra LoRA modules. This makes the %total trainable parameters 
model size unchanged compared to \cite{shi2024dragdiffusion}. %, while still enabling precise and efficient drag-based editing. %Moreover, besides the LoRA for reconstruction finetuning on the given image, we do not incorporate extra LoRA and use the initialization from the reconstruction finetuning on the given single image, thus model parameters are not increased compared with \cite{shi2024dragdiffusion}.

However, we find that optimizing LoRA exclusively for drag-based editing leads to performance degradation, as iterative fine-tuning causes the LoRA-enhanced model to deviate significantly from the original pretrained model. To address this, we propose a dual-objective optimization scheme, combining the drag loss \( L_{\text{drag}} \) with a delta denoising score (DDS) loss \( L_{\text{DDS}} \) \cite{hertz2023delta, arar2024palp}.  Specifically, we first use DragLoRA to predict the clean signal $\hat{z}_0$ based on the feature $z_{35}$ from DDIM inversion. The forward process is then applied, adding noise according to (\ref{eq:eq1}), transforming \( \hat{z}_0 \) into \( \hat{z}_{t'} \), where \( t' \) is a random timestep. Finally, \( \hat{z}_{t'} \) is passed through both the LoRA-enhanced and pretrained models to compute the discrepancy between their noise predictions \( \epsilon^{drag} \) and \( \epsilon^{ori} \). The gradient of the DDS loss is computed as follows: 
\begin{equation}\label{eq:eq5}  
%\frac{\partial{L_{DDS}}}{\partial {\Delta\theta}}
\nabla_{\Delta\theta} L_{\text{DDS}}=(\epsilon^{ori}-\epsilon^{drag})\frac{\partial{\hat{z_0}}}{\partial {\Delta\theta}}
\end{equation}  
Note that this gradient only takes effect through \( \hat{z}_0 \). 
The total loss can be expressed as $L = L_\text{drag} + \lambda_{Mask} L_\text{Mask} + \lambda_{DDS} L_\text{DDS}$. 
While the drag loss aligns handle points with their target positions, the additional gradient ensures consistency with the pretrained model. This dual-objective optimization (DOO) effectively balances precise deformation control with fidelity to the original model, mitigating the instability introduced by unrestricted LoRA optimization.

%To further stabilize the optimization, we introduce a **cyclic denoise-renoise process**. At each iteration, the input feature is denoised from timestep \( t \) to \( t-1 \) using DragLoRA’s predictions, then re-noised back to \( t \) with random perturbations. This cycle propagates handle point adjustments into the latent space, ensuring coherent updates across iterations.  

%By combining online LoRA adaptation, dual-objective optimization, and cyclic feature propagation, DragLoRA achieves precise and efficient drag-based editing, addressing the limitations of static latent optimization and large search spaces in prior work.  

\subsection{Input Latent Feature Adaptation (ILFA)}  
\label{ilfa}
To enhance the stability of motion supervision, we introduce a cyclic denoise-renoise process that adapts the input latent feature for drag-based editing. At each iteration, the input feature \( z_{35}\) at timestep $t=35$ is first denoised to \( t-1 \) using DragLoRA’s predictions, then re-noised back to \( t \) with random Gaussian noise. %following the DDPM schedule as (\ref{eq:eq1}). 
This cycle propagates handle point adjustments into the latent feature space, ensuring coherent updates on LoRA parameters across iterations.  

For denoising, we use the full model, including LoRA, to perform one step of DDIM denoising as defined in (\ref{eq:eq2}). Although DDIM can also be used for re-noising, we find that it produces inferior results compared to the DDPM schedule defined in (\ref{eq:eq1}). Moreover, this scheme is only carried out within the foreground mask, and background region keeps untouched. The proposed ILFA scheme, combined with the dual-objective loss, achieves robust and stable optimization, effectively addressing the challenges of unrestricted LoRA fine-tuning.  
%To further stabilize the optimization in motion supervision, we propose a cyclic denoise-renoise process, which adapts the input feature for drag editing. At each iteration, the input feature is denoised from timestep \( t \) to \( t-1 \) using DragLoRA’s predictions, then re-noised back to \( t \) with random perturbations, where $t=35$. This cycle propagates handle point adjustments into the latent feature space, ensuring coherent updates across iterations. For the denoising, we use the full model including DragLoRA and DDIM in (\ref{eq:eq2}). While the renoising step utilizes randomly sampled Gaussian noise according to the DDPM schedule. In practice, we find that DDIM can be also employed in the renoising step, but achieve inferior results, and the proposed input feature adaptation scheme works well with dual-obejective loss. 

\begin{algorithm}[htb]
   \caption{Drag Updates}
   \label{alg}
\begin{algorithmic}
   \STATE {\bfseries Input:} inverted latent code $z_t$, mask $M$, unet $\theta$, reconstruction lora $\Delta\theta_{rec}$, source points $p$, target points $g$
   \STATE \textbf{Initialize} $\Delta\theta=\Delta\theta_{rec}$, $z_t^0=z_t$, $h=p$, $k=0$
    \STATE $F^0=$ApplyModel$(\theta+\Delta\theta, z_t^0)$
    \WHILE{$k<K$ and $||h-g||_2>l_1$}
   \STATE $F=$ApplyModel$(\theta+\Delta\theta, z_t)$
   \STATE $h,minD=$PointTrack$(F, F^0, p, g, h)$
   \IF{$k>k_{ini}$ and $minD<d_1$ and $||h-n||_{2}<l_2$}
    \WHILE{$minD<d_2$ and $||h-g||_2>l_1$}
   \STATE Adapt $z_t$ 
   \STATE $F=$ApplyModel$(\theta+\Delta\theta, z_t)$
   \STATE $h,minD=$PointTrack$(F, F^0, p, g, h)$
   \ENDWHILE
   \ELSE
   \STATE $d=\frac{g-h}{\|g-h\|_2}$
   \STATE $n=h+d$
   \STATE Optimize $\Delta\theta$ by losses defined as Eq.\eqref{eq:eq3}\eqref{eq:eq4}\eqref{eq:eq5}
   \STATE Adapt $z_t$ 
   \STATE $k+=1$
   \ENDIF
   \ENDWHILE
\end{algorithmic}
\end{algorithm}

\subsection{Adaptive Optimization Scheme with Two Modes}
\label{ass}
The input adaptation scheme works effectively with motion supervision. In some cases, it can drive the handle points toward their targets even without LoRA updates. This is because DragLoRA learns to move the handle points towards the desired direction through previous optimization steps, reducing the need for further LoRA adjustments. This strategy has high efficiency, as it does not need backpropagation.

To balance efficiency and robustness, we propose an adaptive switching scheme (ASS) between two modes, DOO plus ILFA and ILFA-only, based on the quality of point tracking. In the ILFA-only mode, DragLoRA updates the latent feature to guide the handle points to the target positions. When point tracking locates a confident handle point with a small enough $minD$ defined in (\ref{eq:eq7}), %which is similar enough with the feature at source point $\mathbf{p}$ 
and is not far from the temporal target $\mathbf{h+d}$, %LoRA updates are bypassed, as 
ILFA-only is activated. However, in challenging situations such as occlusions or texture ambiguities, point tracking may degrade, requiring further refinement. In these cases, DOO plus ILFA begins so that a gradient-based objective is used to adjust LoRA parameters and stabilize the deformation. This ensures that the handle points remain accurately positioned even in difficult scenarios.
The proposed ASS dynamically toggles between these two modes based on point tracking quality. %When tracking is accurate, input adaptation is used to optimize efficiency. When tracking quality drops, motion supervision is triggered to ensure deformation stability. 
This adaptive scheme makes DragLoRA to efficiently handle a variety of scenarios while maintaining robustness. Details about it are provided in Algorithm \ref{alg}.

\begin{figure}[h]
\vskip 0.1in
\begin{center}
\centerline{\includegraphics[width=(\columnwidth)*9/10]{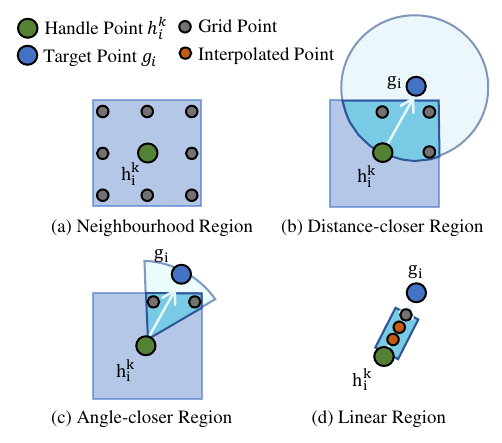}}
\caption{Comparison among different point tracking schemes. (a) A common strategy locates new point in a square neighborhood around current handle point. (b) and (c) reduces the search region using target point, only grid points in the intersection region are considered as candidate. (d) uses a linear line and needs feature interpolation. (b) and (d) are initially proposed in \cite{jiang2024clipdrag} and \cite{ling2024freedrag}, respectively. }
\label{pttrackingcmp}
\end{center}
\vskip -0.1in
\end{figure}

\subsection{Efficient Point Tracking (EPT)}
\label{ept}
To further enhance the efficiency of drag editing, we investigate different point tracking strategies for handle points, as shown in Figure \ref{pttrackingcmp}. We find that DragLoRA can effectively reduce the search region, and it outperforms traditional methods in terms of efficiency. Unlike conventional neighborhood searching, the distance-closer \cite{jiang2024clipdrag} and angle-closer regions constrain the candidate points for handle tracking, preventing unnecessary location reversion. Specifically, the target or current handle point serves as the center of a circle, with their distance defining the radius. Only the grid points within the intersection of the circle (or sector) and the neighboring region are considered as candidates for comparison with the original source point feature. A more aggressive search strategy, proposed in \cite{ling2024freedrag}, follows a straight line from the handle to the target point. In DragLoRA, we choose the distance-closer and angle-closer strategies for point tracking, as the former provides the best performance and the latter is the most efficient. Additionally, to prevent insufficient optimization due to rapid advancement of point coordinates, we determine whether to proceed or retain the previous handle point based on the minD. Further details are provided in \cref{appendix-pt}.
%To further increase the efficiency of drag editing, different point tracking strategy for handle point are investigated, as is shown in Figure \ref{pttrackingcmp}. We find that DragLoRA can effectively reduce the search region, and it even shows better performances in more efficient schemes. Different from traditional neighborhood searching, distance-closer and angle-closer region further constrain candidates for handle point to prevent revert point location. Specifically, the target or current handle point is used as the center of the circle, and the distance from the center to the target is used as the radius. The only grid points in the intersection region between the circle (or sector) and the neighbor region are used as the candidates for comparison with the original source point feature. A more aggressive and straightforward searching region, proposed in \cite{ling2024freedrag}, is along the straight line from handle to the target point. We choose the distance-closer and angle-closer for point tracking in DragLoRA, since the former gives the best performance and the latter is the most efficient one. Details are given in the experiments.

\section{Experiments}

\subsection{Implementation Details}
We use Stable Diffusion 1.5 \cite{rombach2022high} as the
base model. Following DragDiffusion \cite{shi2024dragdiffusion}, we train reconstruction LoRA for each image in 80 steps with a learning rate of 0.0005. Given a total of 50 timesteps, we optimize DragLoRA at $t=35$ with a learning rate of 0.0001. The ranks of both LoRAs are set as 16, and DragLoRA is initialized from RecLoRA. We employ the Adam optimizer, and set $\lambda_{Mask}=0.1, \lambda_{DDS}=50, K=80, k_{ini}=10, l_1=1,l_2=1.4,d_1=1,d_2=1.3$. In terms of EPT, DragLoRA primarily utilizes the distance-closer region. To speed up the dragging process, the angle-closer region is used in DragLoRA-Fast. After drag updates, we apply DragLoRA to all the remaining timesteps to enhance the drag effects. If not specified otherwise, all of our experiments are conducted on a single NVIDIA RTX 4090 GPU.

\subsection{Qualitative Evaluation}
\begin{figure}[h]
% \vskip 0.2in
\begin{center}
\centerline{\includegraphics[width=\columnwidth]{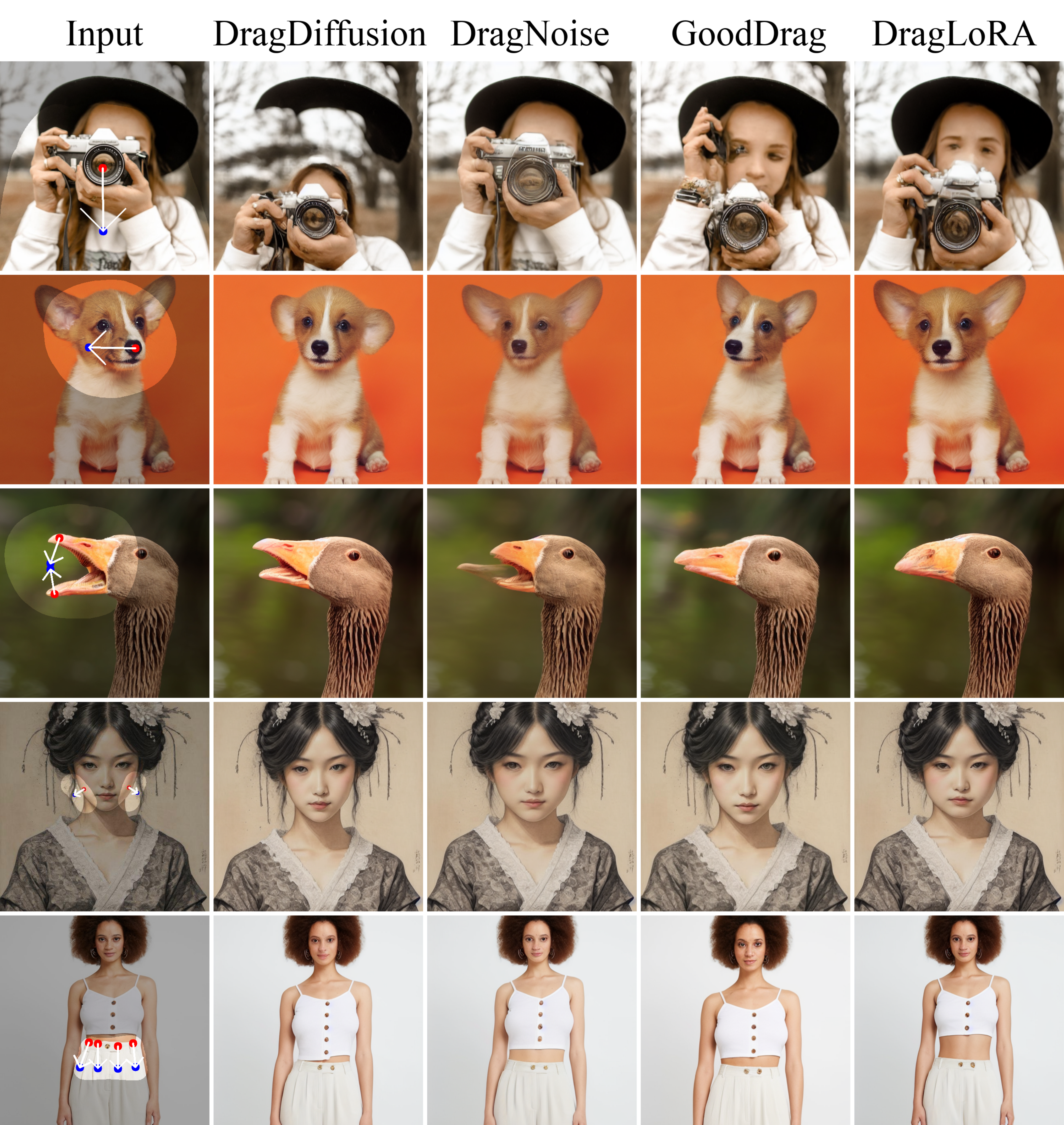}}
\caption{Qualitative comparisons with \cite{shi2024dragdiffusion,liu2024dragnoise,zhang2024gooddrag}. The proposed DragLoRA outperforms existing approaches in both perceptual quality and the accuracy of drag editing.}
\label{visualcmp}
\end{center}
\vskip -0.2in
\end{figure}

To validate the effectiveness of our proposed DragLoRA, we conduct extensive experiments on DragBench \cite{shi2024dragdiffusion}, Drag100 \cite{zhang2024gooddrag}, VITON-HD\cite{choi2021viton} and private collections. We compare DragLoRA qualitatively with existing drag methods: DragDiffusion \cite{shi2024dragdiffusion}, DragNoise \cite{liu2024dragnoise}, GoodDrag \cite{zhang2024gooddrag}.
The visual results are presented in \cref{visualcmp}. Our method and GoodDrag both demonstrate superior editability, while DragDiffusion and DragNoise fail to achieve the target specified by the input annotations, such as not being able to close the duck's mouth (third row). Additionally, compared to GoodDrag, our DragLoRA maintains higher fidelity. For example, after moving the camera, our method generates a more natural-looking face (first row).

\begin{figure}[h]
\vskip 0.2in
\begin{center}
\centerline{\includegraphics[width=\columnwidth]{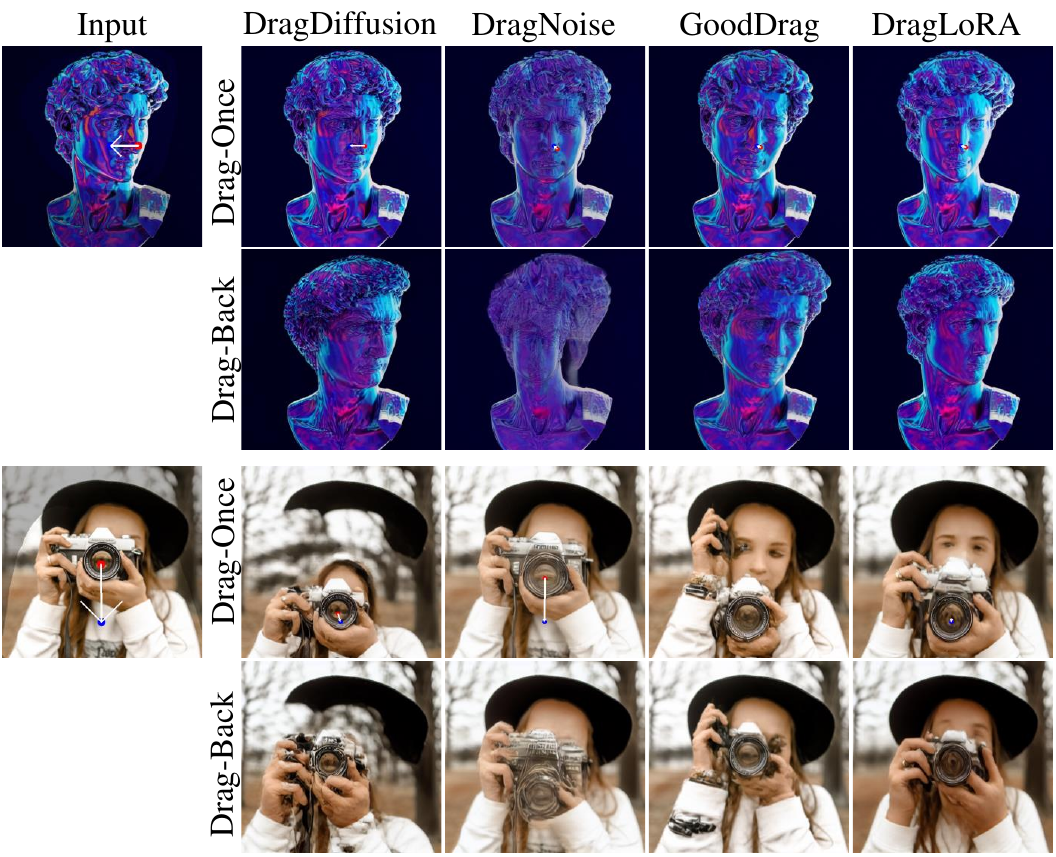}}
\caption{Comparisons in Drag-Back pipeline. Following \cite{ling2024freedrag}, we perform two symmetric drag edits: the first adheres to the input drag annotation and the second reverses it. We focus on the similarity between the drag-back image and the input to validate image fidelity and edit accuracy.}
\label{dragback}
\end{center}
\vskip -0.1in
\end{figure}

\subsection{Quantitative Evaluation}

\begin{table*}[htbp]
\centering
\caption{Quantitative comparisons on DragBench. \textbf{Bold}: best, \underline{underline}: second best, *: not open-sourced. DragLoRA adopts distance-closer EPT while DragLoRA-Fast utilizes angle-closer EPT. Our methods demonstrate state-of-the-art editing quality and the lowest time consumption among optimization-based approaches.}
\label{all_cmp}
\begin{center}
\begin{small}
\begin{sc}
\begin{tabular}{@{}lcccccccc@{}}
\toprule
Methods       & 1-LPIPS $\uparrow$ & MD $\downarrow$                       & m-MD $\downarrow$  & Time(s) $\downarrow$ & Params(M) & RecSteps & DragSteps & Category \\ \midrule
DragDiffusion & 0.88  & 32.13                    & 30.71 & 32.93   & 0.07      & 80       & 80        & Optim    \\
DragNoise     & 0.89  & 35.17                    & 30.66 & 30.53   & 0.33      & 200      & 80        & Optim    \\
StableDrag    & \textbf{0.91}  & 36.46                    & 35.88 & 37.21   & 0.07      & 80       & 80        & Optim    \\
EasyDrag      & \textbf{0.91}      &      38.28           & 37.67 & 44.69   & 0.07      & 0        & 80        & Optim    \\
GoodDrag      & 0.87  & \underline{24.26}                    & \textbf{21.86} & 56.97   & 0.07      & 70       & 210       & Optim    \\
FreeDrag      & \underline{0.90}  & 32.30                    & 30.37 & 51.42   & 0.07      & 200      & 300       & Optim    \\
DragText      & 0.87  & 32.87                    & 29.59 & 42.93   & 0.12      & 80       & 80        & Optim    \\
AdaptiveDrag  & 0.86  & 35.70                    & 32.94 & 77.73   & 0.07      & 80       & 300       & Optim    \\
ClipDrag*     & 0.88  & \multicolumn{1}{l}{32.30} & /     & /       & 0.07      & 160      & 160       & Optim    \\
GDrag* & \textbf{0.91} & 26.49 & / & 152 & 0.08 & 80 & 250 & Optim    \\
DragLoRA      & 0.87  &  \textbf{23.77}    & \underline{22.70} &  \underline{29.84}  & 3.19      & 80       & 80        & Optim    \\ 
DragLoRA-Fast     & 0.88  &  27.55    & 26.90 &  \textbf{22.89}  & 3.19      & 80       & 80        & Optim    \\ 
\midrule
InstantDrag   &    0.88   &   31.51                       &  28.67     &    \textbf{1.44}     &    914       & 0        & 0         & Enc      \\
LightningDrag &  0.89     &  29.10                        &   26.77    &     \underline{1.95}    &     933      & 0        & 0         & Enc      \\ \midrule
SDE-Drag & \textbf{0.91} & 44.48 & 41.53 & 62.74 & 0 & 100 & 0 & TrFree \\
FastDrag      &  0.86     &    31.62                      &   28.29    &   5.13      &      0     &    0      &    0       & TrFree     \\ \bottomrule
\end{tabular}
\end{sc}
\end{small}
\end{center}
\end{table*}

To further compare with more methods \cite{cui2024stabledrag,hou2024easydrag,ling2024freedrag,choi2024dragtext,chen2024adaptivedrag,jiang2024clipdrag,lin2025gdrag,shin2024instantdrag,shi2024lightningdrag,nie2024blessing,zhao2024fastdrag}, we conduct metric evaluations on DragBench. We use publicly available code from other works, except for ClipDrag \cite{jiang2024clipdrag} and GDrag\cite{lin2025gdrag}, which has not open-sourced its code, and categorize all drag methods into three types: optimization-based (Optim), encoder-based(Enc), and training-free (TrFree). We test DragText \cite{choi2024dragtext} based on DragDiffusion \cite{shi2024dragdiffusion}. We evaluate the number of trainable parameters, the steps required to pre-train the reconstruction LoRA, the maximum step set for online optimization, and the associated time. Note that since the reconstruction LoRA can be trained offline and reused for multiple drag editings by various annotations of points, the time spent on this process is not included, which is about 48s per image over 80 steps on one NVIDIA 4090 GPU. As EasyDrag \cite{hou2024easydrag} requires more computational resources, we individually evaluate it on one NVIDIA A40 GPU. 

\textbf{Evaluation Metrics}: Following \cite{shi2024dragdiffusion}, we adopt 1-LPIPS \cite{zhang2018lpips} and MD (Mean Distance) to assess consistency with the original image and editing accuracy, respectively. The former calculates 1 minus feature difference between the original and edited images using AlexNet \cite{krizhevsky2012alex}, while the latter computes coordinate discrepancy between handle points in the edited image and target points using DIFT \cite{tang2023dift}. However, We find that using DIFT to search point coordinates across the entire image may not be accurate, leading to unreliable metrics. Similar to \cite{lu2024regiondrag}, we introduce m-MD (masked-MD), which restricts the DIFT search into the edit region specified by input mask $M$, thereby reducing uncertain errors. While m-MD is numerically lower than MD, it may yield optimistic scores in cases of image distortion, where the mask constraint forces final handle points closer to the target. Thus, the two metrics complement each other. As demonstrated in \cref{all_cmp}, DragLoRA achieves the best MD and competitive m-MD, with enhanced efficiency, compared to the state-of-the-art method GoodDrag. And our DragLoRA-Fast, which adopts angle-closer EPT, demonstrates significant time efficiency among optimization-based approaches while maintaining strong editability.

\textbf{Drag-Back Evaluation}: As the extent of editing increases, 1-LPIPS naturally decreases. Especially in cases of significant edits, it fails to reflect the consistency. Following \cite{ling2024freedrag}, we take a drag-back pipeline to simultaneously measure editability and consistency. After one round of editing, we train a reconstruction LoRA on the edited image, swap the source points and target points, and perform a second round of drag editing. The discrepancy between the second-edited image and the original image is measured by LPIPS and CLIP \cite{radford2021clip}. The CLIP metric calculates the similarity of features extracted by the CLIP image encoder from two images. As shown in \cref{dragback,dragback-ccsd}, lower LPIPS and higher CLIP indicate that both rounds of drag editing effectively preserve the original image information, and also suggest that the first round of editing brings the image sufficiently close to the target. Our DragLoRA achieves better drag-back results both visually and quantitatively.

\begin{table}[h]
\caption{Drag-Back evaluation on DragBench.}
\label{dragback-ccsd}
\vskip 0.1in
\begin{center}
\begin{small}
\begin{sc}
\begin{tabular}{lccc}
\toprule
Methods       & \multicolumn{1}{l}{LPIPS (x10) $\downarrow$} & \multicolumn{1}{l}{CLIP $\uparrow$} \\
\midrule
DragDiffusion & 1.39                      & 0.96                       \\
DragNoise  & 1.64            &    0.94           \\
DragLoRA    & \textbf{1.33}                     & 0.96            \\
GoodDrag  & 1.37             & \textbf{0.97} \\
\bottomrule
\end{tabular}
\end{sc}
\end{small}
\end{center}
\vskip -0.1in
\end{table}

\subsection{Ablation Study}
To systematically evaluate the contributions of different modules to the overall performance, we conduct an ablation study, starting from the baseline and incrementally adding modules. We assess their impact on performance metrics 1-LPIPS and MD. The results are summarized in \cref{ablation}. 

Our baseline starts by optimizing LoRA instead of input latent feature $z_{35}$, based on DragDiffusion. Due to the excessive updates of LoRA, distortive changes occur in the image, resulting in poor performance. Applying DOO effectively stabilizes the drag updates, improving both LPIPS and CLIP. ILFA helps strengthen the edit accuracy by aligning the layout information of input features with LoRA weights. The EPT and ASS are primarily designed to enhance efficiency. The former mitigates the interference from error-tracked points, while the latter makes the training process more adaptive to diverse scenarios, thereby also improving editability. More visible results can be found in \cref{appendix-ablation}.

\begin{table}[h]
\caption{Ablation study on DragLoRA.}
\label{ablation}
\vskip 0.1in
\begin{center}
\begin{small}
\begin{sc}
\begin{tabular}{lcccc}
\toprule
Methods       & \multicolumn{1}{l}{1-LPIPS $\uparrow$} & \multicolumn{1}{l}{CLIP $\uparrow$} & \multicolumn{1}{l}{MD $\downarrow$} & \multicolumn{1}{l}{m-MD $\downarrow$} \\
\midrule
Baseline    & 0.87    & 9.47     & 48.55          &   41.00   \\
+ DOO    & \textbf{0.93}             & \textbf{9.8}             & 36.88    & 34.21              \\
+ ILFA  & 0.88                      & 9.74  & 26.99 &  26.78\\
+ EPT  & 0.88 & 9.74 & 25.45 & 25.24\\
+ ASS  & 0.87  & 9.73  & \textbf{23.77} & \textbf{22.70} \\
\bottomrule
\end{tabular}
\end{sc}
\end{small}
\end{center}
\vskip -0.1in
\end{table}

\section{Conclusion}
We introduce DragLoRA, a novel framework for drag-based image editing that improves precision and efficiency through online optimization of LoRA adapters. By replacing latent feature optimization with dynamic model adaptation, DragLoRA enables finer deformations while preserving semantic fidelity. The dual-objective optimization, combining drag loss and DDS loss, ensures alignment with pretrained diffusion priors, addressing instability from unrestricted LoRA fine-tuning. The cyclic input feature adaptation and adaptive optimization further stabilize motion supervision and boost efficiency. Experiments show DragLoRA outperforms existing methods in both precision and runtime, making it a powerful tool for interactive image editing. Future work will extend this framework to support flexible drag tasks, including region-based drag, incorporating reference images, and various types of drag adaptation.

\textbf{Limitations}.
Our work has several limitations that we leave for the future work. First, considering the quantitative metrics are not direct and accurate, we plan to conduct a user study in the future. Second, there are challenging cases where DragLoRA does not perform optimally, producing edited image with low fidelity, such as moving the camera down to show up the face behind it, which also challenges other methods. We believe advanced generative models can be applied to improve the quality of edited images.
% \section*{Acknowledgements}

% \textbf{Do not} include acknowledgements in the initial version of
% the paper submitted for blind review.

% If a paper is accepted, the final camera-ready version can (and
% usually should) include acknowledgements.  Such acknowledgements
% should be placed at the end of the section, in an unnumbered section
% that does not count towards the paper page limit. Typically, this will 
% include thanks to reviewers who gave useful comments, to colleagues 
% who contributed to the ideas, and to funding agencies and corporate 
% sponsors that provided financial support.
\section*{Acknowledgements}
This work was supported by the Science and Technology Commission of Shanghai Municipality under Grant No.22511105800, 19511120800 and 22DZ2229004, the AI project from the economic and information commission of shanghai (Grant No. 2024-GZL-RGZN-01038), and ECNU Multifunctional Platform for Innovation (001).

\section*{Impact Statement}
This paper presents work whose goal is to advance the field of Machine Learning. There are many potential societal consequences of our work, none which we feel must be specifically highlighted here.

% In the unusual situation where you want a paper to appear in the
% references without citing it in the main text, use \nocite
\nocite{langley00}

\bibliography{draglora}

\begin{thebibliography}{45}
\providecommand{\natexlab}[1]{#1}
\providecommand{\url}[1]{\texttt{#1}}
\expandafter\ifx\csname urlstyle\endcsname\relax
  \providecommand{\doi}[1]{doi: #1}\else
  \providecommand{\doi}{doi: \begingroup \urlstyle{rm}\Url}\fi

\bibitem[Arar et~al.(2024)Arar, Voynov, Hertz, Avrahami, Fruchter, Pritch, Cohen-Or, and Shamir]{arar2024palp}
Arar, M., Voynov, A., Hertz, A., Avrahami, O., Fruchter, S., Pritch, Y., Cohen-Or, D., and Shamir, A.
\newblock Palp: prompt aligned personalization of text-to-image models.
\newblock In \emph{SIGGRAPH Asia 2024 Conference Papers}, pp.\  1--11, 2024.

\bibitem[Cao et~al.(2023)Cao, Wang, Qi, Shan, Qie, and Zheng]{cao2023masactrl}
Cao, M., Wang, X., Qi, Z., Shan, Y., Qie, X., and Zheng, Y.
\newblock Masactrl: Tuning-free mutual self-attention control for consistent image synthesis and editing.
\newblock In \emph{Proceedings of the IEEE/CVF International Conference on Computer Vision}, pp.\  22560--22570, 2023.

\bibitem[Chen et~al.(2024)Chen, Chen, Geng, and Bo]{chen2024adaptivedrag}
Chen, D., Chen, B., Geng, Y., and Bo, L.
\newblock Adaptivedrag: Semantic-driven dragging on diffusion-based image editing.
\newblock \emph{arXiv preprint arXiv:2410.12696}, 2024.

\bibitem[Choi et~al.(2024)Choi, Jeong, Hong, Joo, and Hwang]{choi2024dragtext}
Choi, G., Jeong, T., Hong, S., Joo, J., and Hwang, S.~J.
\newblock Dragtext: Rethinking text embedding in point-based image editing.
\newblock \emph{arXiv preprint arXiv:2407.17843}, 2024.

\bibitem[Choi et~al.(2021)Choi, Park, Lee, and Choo]{choi2021viton}
Choi, S., Park, S., Lee, M., and Choo, J.
\newblock Viton-hd: High-resolution virtual try-on via misalignment-aware normalization.
\newblock In \emph{Proceedings of the IEEE/CVF conference on computer vision and pattern recognition}, pp.\  14131--14140, 2021.

\bibitem[Cui et~al.(2024)Cui, Zhao, Zhang, Cao, Ma, and Wang]{cui2024stabledrag}
Cui, Y., Zhao, X., Zhang, G., Cao, S., Ma, K., and Wang, L.
\newblock Stabledrag: Stable dragging for point-based image editing.
\newblock In \emph{European Conference on Computer Vision (ECCV)}, 2024.

\bibitem[Dhariwal \& Nichol(2021)Dhariwal and Nichol]{dhariwal2021diffusion}
Dhariwal, P. and Nichol, A.
\newblock Diffusion models beat gans on image synthesis.
\newblock \emph{Advances in neural information processing systems}, 34:\penalty0 8780--8794, 2021.

\bibitem[Gal et~al.(2022)Gal, Alaluf, Atzmon, Patashnik, Bermano, Chechik, and Cohen-Or]{gal2022image}
Gal, R., Alaluf, Y., Atzmon, Y., Patashnik, O., Bermano, A.~H., Chechik, G., and Cohen-Or, D.
\newblock An image is worth one word: Personalizing text-to-image generation using textual inversion.
\newblock \emph{arXiv preprint arXiv:2208.01618}, 2022.

\bibitem[Hertz et~al.(2022)Hertz, Mokady, Tenenbaum, Aberman, Pritch, and Cohen-Or]{hertz2022prompt}
Hertz, A., Mokady, R., Tenenbaum, J., Aberman, K., Pritch, Y., and Cohen-Or, D.
\newblock Prompt-to-prompt image editing with cross attention control.
\newblock \emph{arXiv preprint arXiv:2208.01626}, 2022.

\bibitem[Hertz et~al.(2023)Hertz, Aberman, and Cohen-Or]{hertz2023delta}
Hertz, A., Aberman, K., and Cohen-Or, D.
\newblock Delta denoising score.
\newblock In \emph{Proceedings of the IEEE/CVF International Conference on Computer Vision}, pp.\  2328--2337, 2023.

\bibitem[Ho et~al.(2020)Ho, Jain, and Abbeel]{ho2020denoising}
Ho, J., Jain, A., and Abbeel, P.
\newblock Denoising diffusion probabilistic models.
\newblock \emph{Advances in neural information processing systems}, 33:\penalty0 6840--6851, 2020.

\bibitem[Hou et~al.(2024)Hou, Liu, Zhang, Liu, Liu, and You]{hou2024easydrag}
Hou, X., Liu, B., Zhang, Y., Liu, J., Liu, Y., and You, H.
\newblock Easydrag: Efficient point-based manipulation on diffusion models.
\newblock In \emph{Proceedings of the IEEE/CVF Conference on Computer Vision and Pattern Recognition}, pp.\  8404--8413, 2024.

\bibitem[Jiang et~al.(2024)Jiang, Wang, and Chen]{jiang2024clipdrag}
Jiang, Z., Wang, Z., and Chen, L.
\newblock Combing text-based and drag-based editing for precise and flexible image editing.
\newblock \emph{arXiv preprint arXiv:2410.03097}, 2024.

\bibitem[Kawar et~al.(2023)Kawar, Zada, Lang, Tov, Chang, Dekel, Mosseri, and Irani]{kawar2023imagic}
Kawar, B., Zada, S., Lang, O., Tov, O., Chang, H., Dekel, T., Mosseri, I., and Irani, M.
\newblock Imagic: Text-based real image editing with diffusion models.
\newblock In \emph{Proceedings of the IEEE/CVF Conference on Computer Vision and Pattern Recognition}, pp.\  6007--6017, 2023.

\bibitem[Krizhevsky et~al.(2012)Krizhevsky, Sutskever, and Hinton]{krizhevsky2012alex}
Krizhevsky, A., Sutskever, I., and Hinton, G.~E.
\newblock Imagenet classification with deep convolutional neural networks.
\newblock \emph{Advances in neural information processing systems}, 25, 2012.

\bibitem[Kumari et~al.(2023)Kumari, Zhang, Zhang, Shechtman, and Zhu]{kumari2023multi}
Kumari, N., Zhang, B., Zhang, R., Shechtman, E., and Zhu, J.-Y.
\newblock Multi-concept customization of text-to-image diffusion.
\newblock In \emph{Proceedings of the IEEE/CVF Conference on Computer Vision and Pattern Recognition}, pp.\  1931--1941, 2023.

\bibitem[Lin et~al.(2025)Lin, Li, Cheng, Yan, and Liang]{lin2025gdrag}
Lin, X., Li, H., Cheng, Y., Yan, Y., and Liang, X.
\newblock Gdrag: Towards general-purpose interactive editing with anti-ambiguity point diffusion.
\newblock In \emph{The Thirteenth International Conference on Learning Representations}, 2025.

\bibitem[Ling et~al.(2024)Ling, Chen, Zhang, Chen, Jin, and Zheng]{ling2024freedrag}
Ling, P., Chen, L., Zhang, P., Chen, H., Jin, Y., and Zheng, J.
\newblock Freedrag: Feature dragging for reliable point-based image editing.
\newblock In \emph{Proceedings of the IEEE/CVF Conference on Computer Vision and Pattern Recognition}, pp.\  6860--6870, 2024.

\bibitem[Liu et~al.(2024)Liu, Xu, Yang, Zeng, and He]{liu2024dragnoise}
Liu, H., Xu, C., Yang, Y., Zeng, L., and He, S.
\newblock Drag your noise: Interactive point-based editing via diffusion semantic propagation.
\newblock In \emph{Proceedings of the IEEE/CVF Conference on Computer Vision and Pattern Recognition}, pp.\  6743--6752, 2024.

\bibitem[Lu et~al.(2024)Lu, Li, and Han]{lu2024regiondrag}
Lu, J., Li, X., and Han, K.
\newblock Regiondrag: Fast region-based image editing with diffusion models.
\newblock In \emph{European Conference on Computer Vision (ECCV)}, 2024.

\bibitem[Meng et~al.(2022)Meng, He, Song, Song, Wu, Zhu, and Ermon]{meng2022sdedit}
Meng, C., He, Y., Song, Y., Song, J., Wu, J., Zhu, J.-Y., and Ermon, S.
\newblock {SDE}dit: Guided image synthesis and editing with stochastic differential equations.
\newblock In \emph{International Conference on Learning Representations}, 2022.

\bibitem[Miyake et~al.(2023)Miyake, Iohara, Saito, and Tanaka]{miyake2023negative}
Miyake, D., Iohara, A., Saito, Y., and Tanaka, T.
\newblock Negative-prompt inversion: Fast image inversion for editing with text-guided diffusion models.
\newblock \emph{arXiv preprint arXiv:2305.16807}, 2023.

\bibitem[Mokady et~al.(2023)Mokady, Hertz, Aberman, Pritch, and Cohen-Or]{mokady2023null}
Mokady, R., Hertz, A., Aberman, K., Pritch, Y., and Cohen-Or, D.
\newblock Null-text inversion for editing real images using guided diffusion models.
\newblock In \emph{Proceedings of the IEEE/CVF Conference on Computer Vision and Pattern Recognition}, pp.\  6038--6047, 2023.

\bibitem[Mou et~al.(2023)Mou, Wang, Song, Shan, and Zhang]{mou2023dragondiffusion}
Mou, C., Wang, X., Song, J., Shan, Y., and Zhang, J.
\newblock Dragondiffusion: Enabling drag-style manipulation on diffusion models.
\newblock \emph{arXiv preprint arXiv:2307.02421}, 2023.

\bibitem[Mou et~al.(2024)Mou, Wang, Song, Shan, and Zhang]{mou2024diffeditor}
Mou, C., Wang, X., Song, J., Shan, Y., and Zhang, J.
\newblock Diffeditor: Boosting accuracy and flexibility on diffusion-based image editing.
\newblock In \emph{Proceedings of the IEEE/CVF Conference on Computer Vision and Pattern Recognition}, pp.\  8488--8497, 2024.

\bibitem[Nie et~al.(2024)Nie, Guo, Lu, Zhou, Zheng, and Li]{nie2024blessing}
Nie, S., Guo, H.~A., Lu, C., Zhou, Y., Zheng, C., and Li, C.
\newblock The blessing of randomness: Sde beats ode in general diffusion-based image editing.
\newblock In \emph{International Conference on Learning Representations}, 2024.

\bibitem[Pan et~al.(2023)Pan, Tewari, Leimk{\"u}hler, Liu, Meka, and Theobalt]{pan2023drag}
Pan, X., Tewari, A., Leimk{\"u}hler, T., Liu, L., Meka, A., and Theobalt, C.
\newblock Drag your gan: Interactive point-based manipulation on the generative image manifold.
\newblock In \emph{ACM SIGGRAPH 2023 Conference Proceedings}, pp.\  1--11, 2023.

\bibitem[Poole et~al.(2022)Poole, Jain, Barron, and Mildenhall]{poole2022sds}
Poole, B., Jain, A., Barron, J.~T., and Mildenhall, B.
\newblock Dreamfusion: Text-to-3d using 2d diffusion.
\newblock \emph{arXiv preprint arXiv:2209.14988}, 2022.

\bibitem[Radford et~al.(2021)Radford, Kim, Hallacy, Ramesh, Goh, Agarwal, Sastry, Askell, Mishkin, Clark, et~al.]{radford2021clip}
Radford, A., Kim, J.~W., Hallacy, C., Ramesh, A., Goh, G., Agarwal, S., Sastry, G., Askell, A., Mishkin, P., Clark, J., et~al.
\newblock Learning transferable visual models from natural language supervision.
\newblock In \emph{International conference on machine learning}, pp.\  8748--8763. PMLR, 2021.

\bibitem[Rombach et~al.(2022)Rombach, Blattmann, Lorenz, Esser, and Ommer]{rombach2022high}
Rombach, R., Blattmann, A., Lorenz, D., Esser, P., and Ommer, B.
\newblock High-resolution image synthesis with latent diffusion models.
\newblock In \emph{Proceedings of the IEEE/CVF conference on computer vision and pattern recognition}, pp.\  10684--10695, 2022.

\bibitem[Ruiz et~al.(2023)Ruiz, Li, Jampani, Pritch, Rubinstein, and Aberman]{ruiz2023dreambooth}
Ruiz, N., Li, Y., Jampani, V., Pritch, Y., Rubinstein, M., and Aberman, K.
\newblock Dreambooth: Fine tuning text-to-image diffusion models for subject-driven generation.
\newblock In \emph{Proceedings of the IEEE/CVF conference on computer vision and pattern recognition}, pp.\  22500--22510, 2023.

\bibitem[Shi et~al.(2024{\natexlab{a}})Shi, Liew, Yan, Tan, and Feng]{shi2024lightningdrag}
Shi, Y., Liew, J.~H., Yan, H., Tan, V. Y.~F., and Feng, J.
\newblock Lightningdrag: Lightning fast and accurate drag-based image editing emerging from videos.
\newblock \emph{arXiv preprint arXiv:2405.13722}, 2024{\natexlab{a}}.

\bibitem[Shi et~al.(2024{\natexlab{b}})Shi, Xue, Liew, Pan, Yan, Zhang, Tan, and Bai]{shi2024dragdiffusion}
Shi, Y., Xue, C., Liew, J.~H., Pan, J., Yan, H., Zhang, W., Tan, V.~Y., and Bai, S.
\newblock Dragdiffusion: Harnessing diffusion models for interactive point-based image editing.
\newblock In \emph{Proceedings of the IEEE/CVF Conference on Computer Vision and Pattern Recognition}, pp.\  8839--8849, 2024{\natexlab{b}}.

\bibitem[Shin et~al.(2024)Shin, Choi, and Park]{shin2024instantdrag}
Shin, J., Choi, D., and Park, J.
\newblock Instantdrag: Improving interactivity in drag-based image editing.
\newblock In \emph{SIGGRAPH Asia 2024 Conference Papers}, pp.\  1--10, 2024.

\bibitem[Song et~al.(2020{\natexlab{a}})Song, Meng, and Ermon]{song2020denoising}
Song, J., Meng, C., and Ermon, S.
\newblock Denoising diffusion implicit models.
\newblock \emph{arXiv preprint arXiv:2010.02502}, 2020{\natexlab{a}}.

\bibitem[Song et~al.(2020{\natexlab{b}})Song, Sohl-Dickstein, Kingma, Kumar, Ermon, and Poole]{song2020score}
Song, Y., Sohl-Dickstein, J., Kingma, D.~P., Kumar, A., Ermon, S., and Poole, B.
\newblock Score-based generative modeling through stochastic differential equations.
\newblock \emph{arXiv preprint arXiv:2011.13456}, 2020{\natexlab{b}}.

\bibitem[Tang et~al.(2023)Tang, Jia, Wang, Phoo, and Hariharan]{tang2023dift}
Tang, L., Jia, M., Wang, Q., Phoo, C.~P., and Hariharan, B.
\newblock Emergent correspondence from image diffusion.
\newblock \emph{Advances in Neural Information Processing Systems}, 36:\penalty0 1363--1389, 2023.

\bibitem[Tumanyan et~al.(2023)Tumanyan, Geyer, Bagon, and Dekel]{tumanyan2023plug}
Tumanyan, N., Geyer, M., Bagon, S., and Dekel, T.
\newblock Plug-and-play diffusion features for text-driven image-to-image translation.
\newblock In \emph{Proceedings of the IEEE/CVF Conference on Computer Vision and Pattern Recognition}, pp.\  1921--1930, 2023.

\bibitem[Valevski et~al.(2023)Valevski, Kalman, Molad, Segalis, Matias, and Leviathan]{valevski2023unitune}
Valevski, D., Kalman, M., Molad, E., Segalis, E., Matias, Y., and Leviathan, Y.
\newblock Unitune: Text-driven image editing by fine tuning a diffusion model on a single image.
\newblock \emph{ACM Transactions on Graphics (TOG)}, 42\penalty0 (4):\penalty0 1--10, 2023.

\bibitem[Wei et~al.(2023)Wei, Zhang, Ji, Bai, Zhang, and Zuo]{wei2023elite}
Wei, Y., Zhang, Y., Ji, Z., Bai, J., Zhang, L., and Zuo, W.
\newblock Elite: Encoding visual concepts into textual embeddings for customized text-to-image generation.
\newblock In \emph{Proceedings of the IEEE/CVF International Conference on Computer Vision}, pp.\  15943--15953, 2023.

\bibitem[Ye et~al.(2023)Ye, Zhang, Liu, Han, and Yang]{ye2023ip}
Ye, H., Zhang, J., Liu, S., Han, X., and Yang, W.
\newblock Ip-adapter: Text compatible image prompt adapter for text-to-image diffusion models.
\newblock \emph{arXiv preprint arXiv:2308.06721}, 2023.

\bibitem[Zhang et~al.(2023)Zhang, Rao, and Agrawala]{zhang2023adding}
Zhang, L., Rao, A., and Agrawala, M.
\newblock Adding conditional control to text-to-image diffusion models.
\newblock In \emph{Proceedings of the IEEE/CVF International Conference on Computer Vision}, pp.\  3836--3847, 2023.

\bibitem[Zhang et~al.(2018)Zhang, Isola, Efros, Shechtman, and Wang]{zhang2018lpips}
Zhang, R., Isola, P., Efros, A.~A., Shechtman, E., and Wang, O.
\newblock The unreasonable effectiveness of deep features as a perceptual metric.
\newblock In \emph{Proceedings of the IEEE conference on computer vision and pattern recognition}, pp.\  586--595, 2018.

\bibitem[Zhang et~al.(2024)Zhang, Liu, Chen, and Xu]{zhang2024gooddrag}
Zhang, Z., Liu, H., Chen, J., and Xu, X.
\newblock Gooddrag: Towards good practices for drag editing with diffusion models.
\newblock \emph{arXiv preprint arXiv:2404.07206}, 2024.

\bibitem[Zhao et~al.(2024)Zhao, Guan, Fan, Xu, Lin, Pan, and Feng]{zhao2024fastdrag}
Zhao, X., Guan, J., Fan, C., Xu, D., Lin, Y., Pan, H., and Feng, P.
\newblock Fastdrag: Manipulate anything in one step, 2024.

\end{thebibliography}
\bibliographystyle{icml2025}

%%%%%%%%%%%%%%%%%%%%%%%%%%%%%%%%%%%%%%%%%%%%%%%%%%%%%%%%%%%%%%%%%%%%%%%%%%%%%%%
%%%%%%%%%%%%%%%%%%%%%%%%%%%%%%%%%%%%%%%%%%%%%%%%%%%%%%%%%%%%%%%%%%%%%%%%%%%%%%%
% APPENDIX
%%%%%%%%%%%%%%%%%%%%%%%%%%%%%%%%%%%%%%%%%%%%%%%%%%%%%%%%%%%%%%%%%%%%%%%%%%%%%%%
%%%%%%%%%%%%%%%%%%%%%%%%%%%%%%%%%%%%%%%%%%%%%%%%%%%%%%%%%%%%%%%%%%%%%%%%%%%%%%%
\newpage
\appendix
\onecolumn
\section{Additional Results}
\begin{figure}[!t]
\vskip 0.1in
\begin{center}
\centerline{\includegraphics[width=(\columnwidth)]{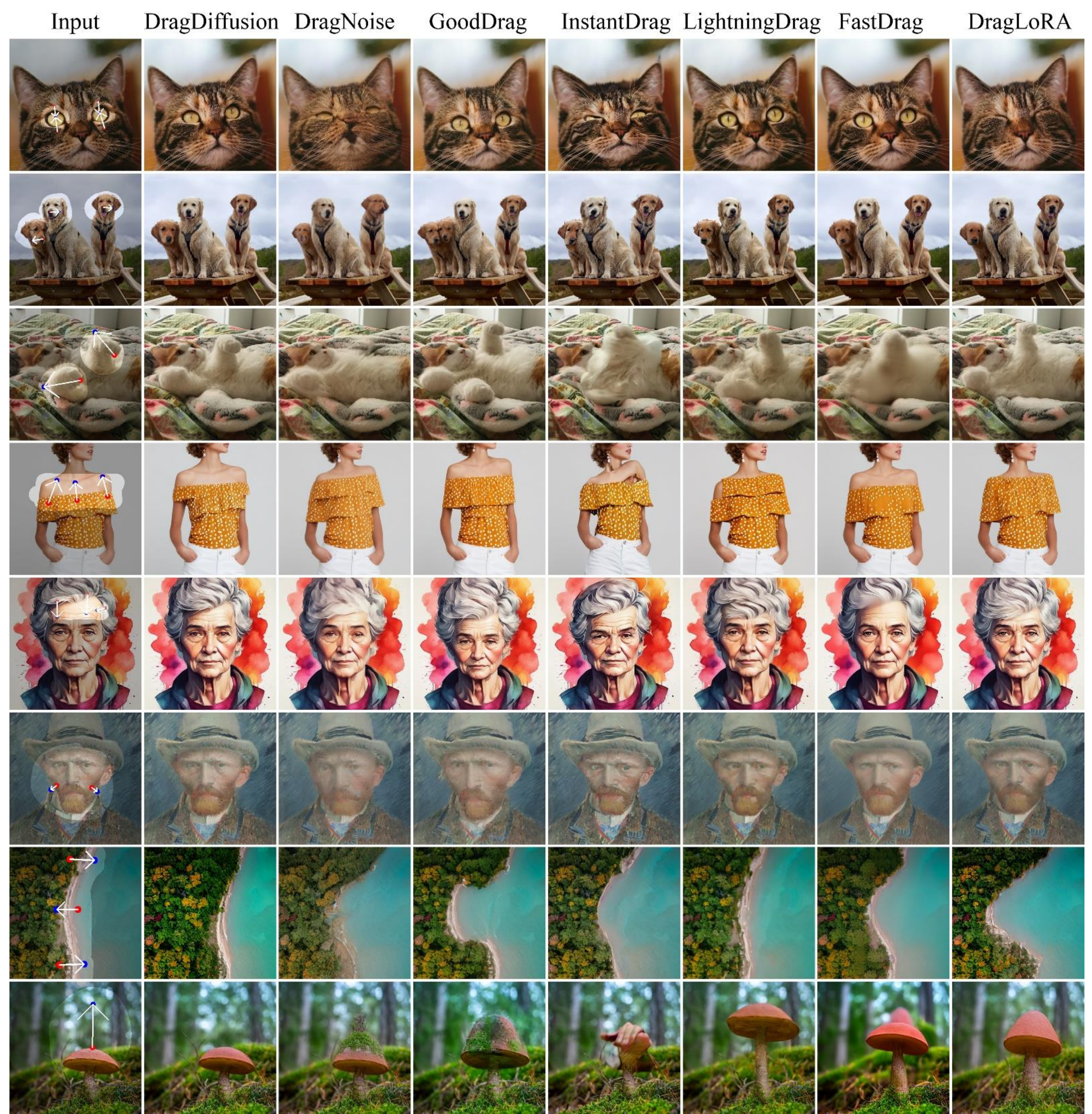}}
\caption{Qualitative comparisons with more methods \cite{shi2024dragdiffusion,liu2024dragnoise,zhang2024gooddrag,shin2024instantdrag,shi2024lightningdrag,zhao2024fastdrag}.
}
\label{appendix-cmp}
\end{center}
\vskip -0.25in
\end{figure}

\begin{figure}[!h]
% \vskip 0.2in
\begin{center}
\centerline{\includegraphics[width=(\columnwidth)*10/10]{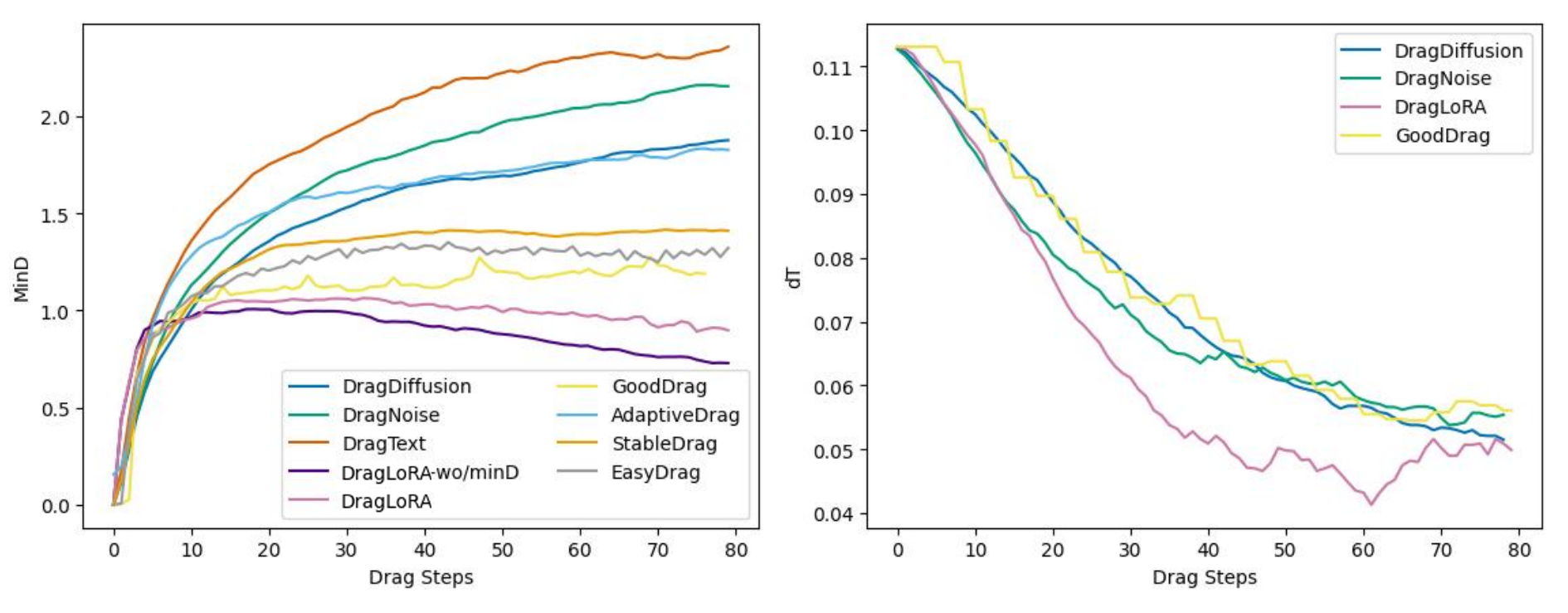}}
\caption{Comparisons of \( minD \) and \( dT \) across different methods. For each benchmark image, we record \( minD \) and \( dT \) from point tracking after each optimization step and compute the average curve across all images. DragLoRA achieves the lowest \( minD \) and \( dT \), indicating superior dragged results.%Comparisons of $minD$ in different methods of point tracking. For each image in the benchmark, we record its $minD$ of point tracking after every optimization of motion supervision, and compute the average curve. DragLoRA achieves the lowest metric.
}
\label{minD}
\end{center}
\vskip -0.2in
\end{figure}

More qualitative comparison results are given in \cref{appendix-cmp}. It can be observed that DragLoRA achieves the most accurate drag-edit results. As evidenced by \cref{minD}, the minD of DragLoRA averaged on 205 images from DragBench reaches the lowest through 80 steps, which demonstrates the stability and reliability of our proposed optimization process. Notably, minD-related strategies (ASS and EPT) are not designed to decrease minD, and minD curve comparisons between DragLoRA and other methods are fair. When ASS and EPT are ablated (DragLoRA-wo/minD, dark purple), the observed minD is even lower than DragLoRA.
To further evaluate the efficiency of point displacement, we compute the Euclidean distance between handle points and target points $d(\mathbf{h}_i^{k+1},\mathbf{g}_i)$ at each drag step, and report the value averaged over all points and images, which is noted as dT. Results in \cref{minD} confirm that DragLoRA drives handle points to target positions more efficiently than various existing methods. 

\section{More discussion}
In the main paper, we have studied the effects of Dual-Object Optimization (DOO), Input Latent Feature Adaptation (ILFA), Adaptive Switching Scheme (ASS) and Efficient Point Tracking (EPT). In this section, we provide more analysis on DOO, ILFA and details bout EPT. Furthermore, we provide a qualitative ablation study as a supplement to the quantitative results.

\subsection{Dual-Object Optimization}
In this section, we analyze the difference between the loss introduced in \cite{hertz2023delta} and our $L_\text{DDS}$. 

In the original DDS paper, $\nabla_{\theta} L_{\text{DDS}}=(\epsilon^{edit}-\epsilon^{ori})\frac{\partial{z_0}}{\partial {\theta}}$ is applied to guide the image towards the semantic editing direction, which points from the original text to the target text. In our work, however, we employ $L_\text{drag}$ for the editing operation and use $\nabla_{\Delta\theta} L_{\text{DDS}}=(\epsilon^{ori}-\epsilon^{drag})\frac{\partial{\hat{z_0}}}{\partial {\Delta\theta}}$ as a regularization term. This regularization constrains the edited model to remain close to the original model’s generative capability, thereby preventing excessive updates. In essence, our application of the DDS loss is intentionally reversed relative to the original purpose. Therefore, the backpropagated gradient in our method carries an additional negative sign compared to the original DDS loss function.
\subsection{Input Latent Feature Adaptation}
In this section, we analyze the necessity and theoretical feasibility of ILFA discussed in \cref{ilfa}, building a connection with Score Distillation Sampling (SDS) \cite{poole2022sds} and delta denoising score (DDS) \cite{hertz2023delta}. Furthermore, we apply ILFA to DragNoise \cite{liu2024dragnoise} to validate its generalization.

\begin{figure}[!h]
% \vskip 0.2in
\begin{center}
\centerline{\includegraphics[width=(\columnwidth)*6/10]{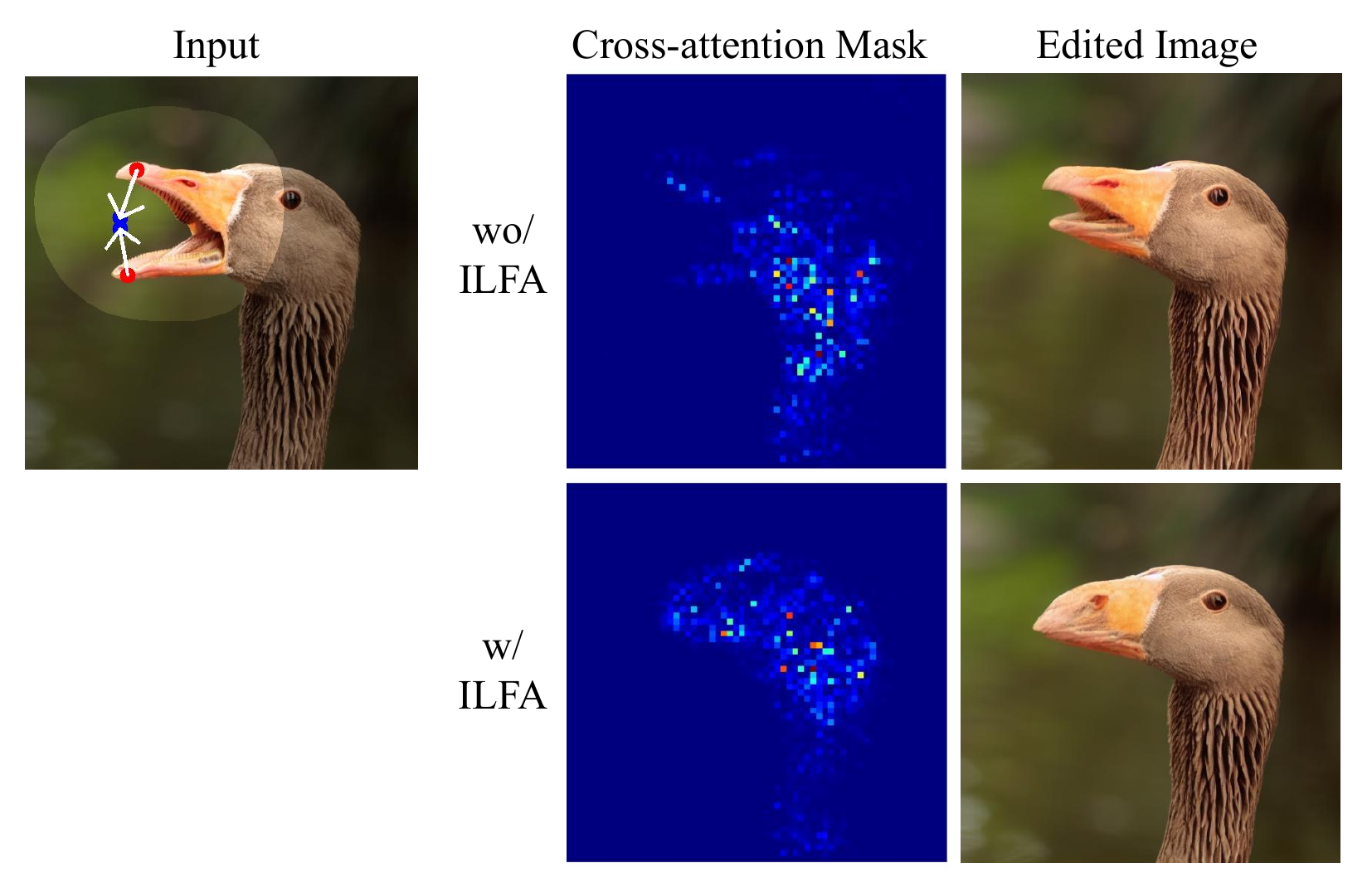}}
\caption{Visualization of cross-attention map to reveal layout conflicts between the fixed input latent feature and DragLoRA. ILFA mitigates these conflicts and produces more precise editing results.
}
\label{appendix-attn}
\end{center}
\vskip -0.2in
\end{figure}

\textbf{Necessity}.
Maintaining a fixed input feature $z_{35}$ while solely optimizing DragLoRA preserves the initial spatial layout. This constraint forces the LoRA to generate the growing displacements as handle points approach their targets, creating a fundamental conflict with the small-step motion supervision paradigm. As visualized in the cross-attention maps (\cref{appendix-attn}), this mismatch manifests as unclear outline of duck's beak and suboptimal editing results. To resolve this instability, we propose Input Latent Feature Adaptation (ILFA), which dynamically aligns $z_{35}$ with the LoRA-learned motion through a denoise-renoise update strategy.

\textbf{Formula Analysis}.
In ILFA, we denoise the input latent feature $z_{t},t=35$ to $z_{t-1}$ by \cref{eq:eq2} and then renoise it to obtain the new $z_{t}$ by \cref{eq:eq1}. We can combine the two equations and rewrite them into a single form as:
\begin{equation}\small\label{eq:eq-appendix} 
\begin{split}
z_{t}&=\sqrt{\alpha_t} \, \left( \sqrt{\bar{\alpha}_{t-1}} \left( \frac{z_t - \sqrt{1 - \bar{\alpha}_t} \, \epsilon_{\theta+\Delta\theta}(z_t, t)}{\sqrt{\bar{\alpha}_t}} \right) + \sqrt{1 - \bar{\alpha}_{t-1}} \cdot \epsilon_{\theta+\Delta\theta}(z_t, t) \right) + \sqrt{1-\alpha_t}\cdot\epsilon_t \\
&= z_t + \sqrt{1-\alpha_t}\cdot\epsilon_t - (\sqrt{1-\bar{\alpha}_t} - \sqrt{\alpha_t - \bar{\alpha}_t})\cdot\epsilon_{\theta+\Delta\theta}(z_t, t)
\end{split}
\end{equation}
Essentially, ILFA involves taking the weighted difference between the DragLoRA prediction and random noise, and then using this difference as the editing direction to guide input feature. This is similar to SDS, which takes the difference as grad to optimize the input feature. If we renoise $z_{t-1}$ with DDIM inversion as \cref{eq:eq3}, the guidance form looks like DDS.
As shown in \cref{cmp-ilfa}, the SDS form works better in DragLoRA. We attribute this to the fact that random noise is more capable of altering the inherent layout information and creating suitable information that needs to be filled in.
% Please add the following required packages to your document preamble:
% \usepackage{booktabs}
\begin{table}[h]
\caption{Comparison on ILFA forms.}
\label{cmp-ilfa}
\vskip 0.1in
\begin{center}
\begin{small}
\begin{sc}
\begin{tabular}{lcc}
\toprule
Methods                      & 1-LPIPS                  & MD                        \\ \midrule
\multicolumn{1}{c}{ILFA-DDS} & \multicolumn{1}{c}{0.91} & \multicolumn{1}{c}{29.33} \\
ILFA-SDS                     & 0.87                     & 23.77                     \\ \bottomrule
\end{tabular}
\vskip 0.1in
\end{sc}
\end{small}
\end{center}
\end{table}

\begin{figure}[!h]
% \vskip 0.2in
\begin{center}
\centerline{\includegraphics[width=(\columnwidth)*9/10]{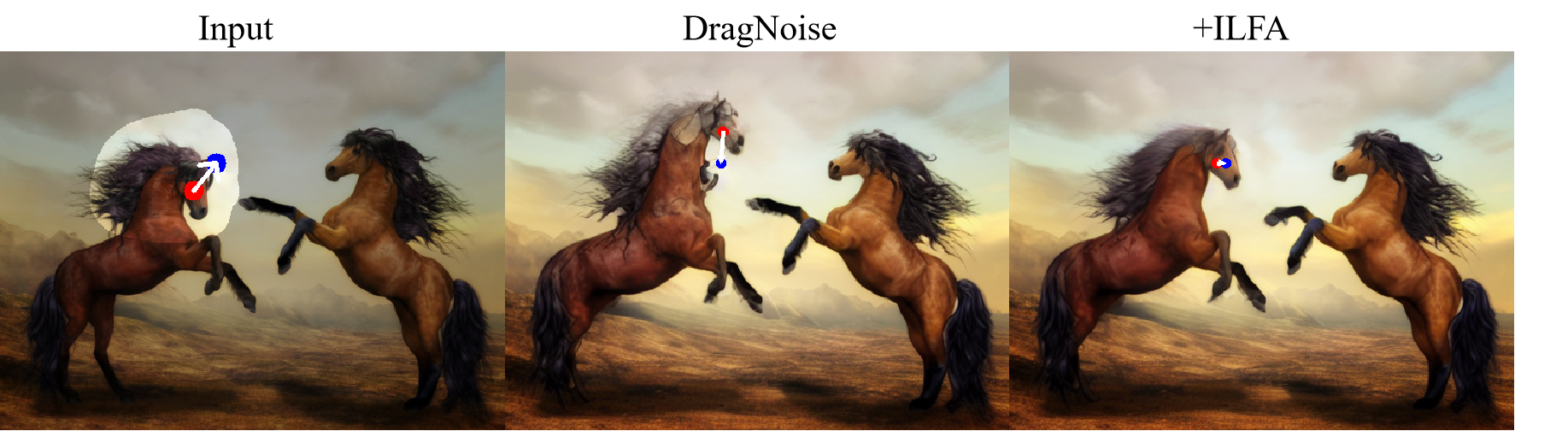}}
\caption{Comparison on DragNoise-based ILFA.
}
\label{dragnoise}
\end{center}
\vskip -0.2in
\end{figure}

\textbf{Generality}.
ILFA is portable and can be applied to other methods like DragNoise, which optimizes the intermediate feature of Unet. As shown in \cref{dragnoise}, ILFA improves the editability and reduces the ambiguity caused by layout conflicts between the input and the intermediate layer.

\subsection{Efficient Point Tracking}
\label{appendix-pt}
As introduced in \cref{ept}, the common neighborhood region is so large that the tracked points could be misled by ambiguous points, causing dragging to be stuck. To improve efficiency and eliminate error-prone reverse-direction points, we tested distance-closer \cite{jiang2024clipdrag}, angle-closer and linear \cite{ling2024freedrag} region on DragLoRA, with a decreasing number of candidate points. The stricter the constraints on the candidates, the more likely the tracked points are to move towards the targets, which may ultimately lead to under-optimization. Therefore, we introduce an additional confidence check. Specifically, when minD $>d_2$, we determine that the tracked point cannot accurately reflect the current editing state, so we retain the previous point coordinates. As shown in \cref{ablation_pt}, the Distance-closer method achieves the best drag edits while the other two cost less time. Since we combine DragLoRA with linear point searching in the FreeDrag way, we also compare it with FreeDrag and find that DragLoRA surpasses FreeDrag in every aspect.

\begin{table}[h]
\caption{Comparison on strategies in EPT.}
\label{ablation_pt}
\vskip 0.1in
\begin{center}
\begin{small}
\begin{sc}
\begin{tabular}{lcccc}
\toprule
Methods       & \multicolumn{1}{l}{1-LPIPS $\uparrow$} & \multicolumn{1}{l}{MD $\downarrow$}  & \multicolumn{1}{l}{m-MD $\downarrow$}  & \multicolumn{1}{l}{Time(s) $\downarrow$}\\
\midrule
Distance-closer    & 0.87   & \textbf{23.77} & \textbf{22.70} & 29.84               \\
Angle-closer    & 0.88 & 27.55 & 26.90  & 22.89             \\
Linear  & \textbf{0.90}   & 30.61 &  30.21	&	\textbf{20.79}
\\
\midrule
FreeDrag & \textbf{0.90} & 32.30 & 30.37 & 51.42\\
\bottomrule
\end{tabular}
\end{sc}
\end{small}
\end{center}
\vskip -0.1in
\end{table}

\subsection{Ablation Study}
\label{appendix-ablation}
As shown in \cref{appendix-ablation-visual}, simply optimizing LoRA accroding to the strategies of DragDiffusion leads to unexpected extreme edits. We increase the image fidelity by applying DOO to restrict the parameters of DragLoRA from deviating largely. As we successively incorporate ILFA, ASS and EPT, DragLoRA significantly enhances the degree of editing, and these significant edits are stable and reliable.
\begin{figure}[!h]
% \vskip 0.2in
\begin{center}
\centerline{\includegraphics[width=(\columnwidth)*8/10]{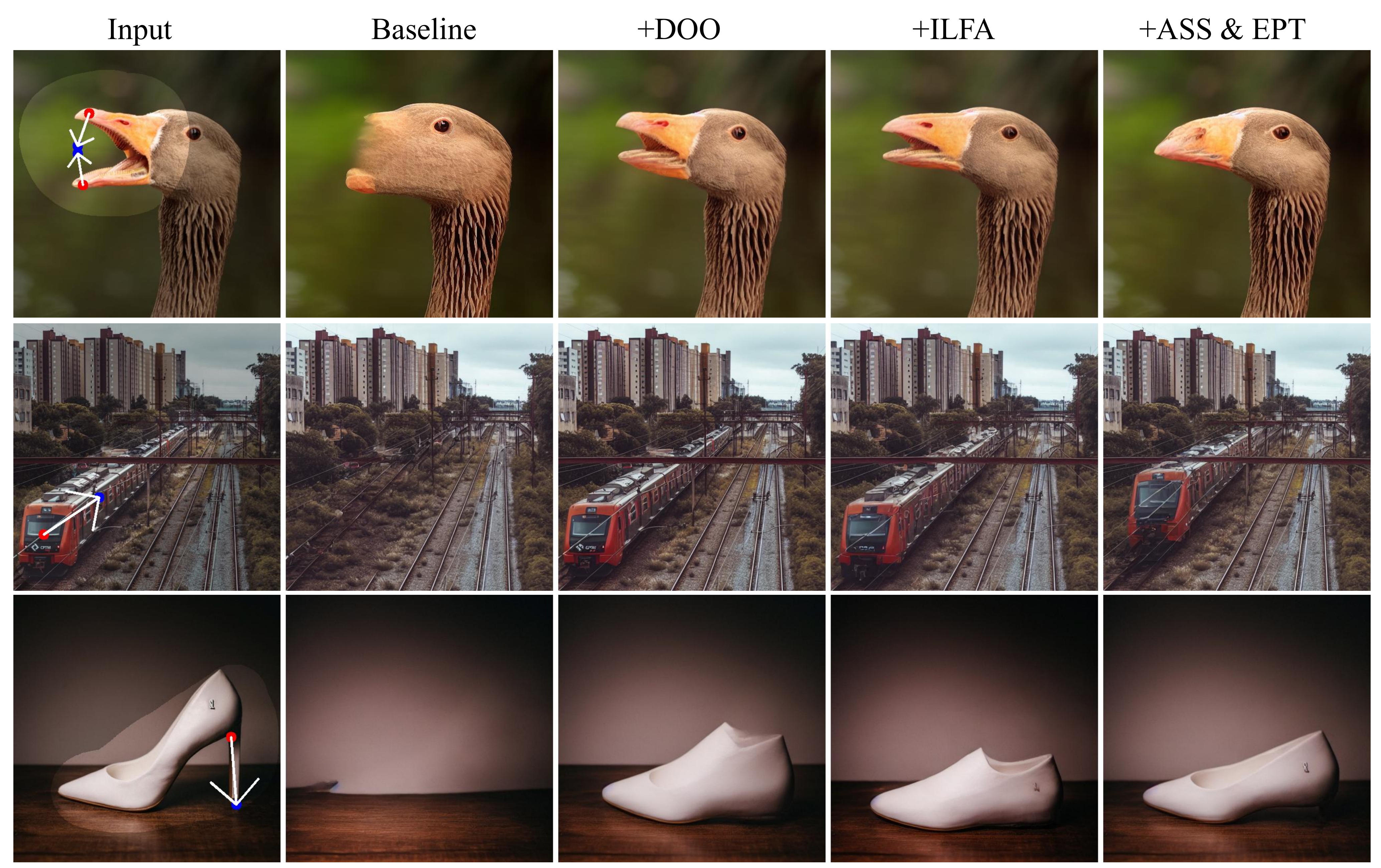}}
\caption{Visual ablation study.
}
\label{appendix-ablation-visual}
\end{center}
\vskip -0.2in
\end{figure}

%%%%%%%%%%%%%%%%%%%%%%%%%%%%%%%%%%%%%%%%%%%%%%%%%%%%%%%%%%%%%%%%%%%%%%%%%%%%%%%
%%%%%%%%%%%%%%%%%%%%%%%%%%%%%%%%%%%%%%%%%%%%%%%%%%%%%%%%%%%%%%%%%%%%%%%%%%%%%%%

\end{document}